\documentclass[10pt,twocolumn,letterpaper]{article}

\usepackage{iccv}
\usepackage{times}
\usepackage{epsfig}
\usepackage{graphicx}
\usepackage{amsmath}
\usepackage{amssymb}

\usepackage{threeparttable}
\usepackage{array}
\usepackage{multirow}
\usepackage{authblk}
\newcolumntype{C}[1]{>{\centering\let\newline\\\arraybackslash\hspace{0pt}}m{#1}}

\usepackage[pagebackref=true,breaklinks=true,letterpaper=true,colorlinks,bookmarks=false]{hyperref}

\iccvfinalcopy % *** Uncomment this line for the final submission

\def\iccvPaperID{6668} % *** Enter the ICCV Paper ID here

\begin{document}

%%%%%%%%% TITLE
\title{Evaluating Robustness of Deep Image Super-Resolution\\Against Adversarial Attacks}

\author[1]{Jun-Ho Choi}
\author[2]{Huan Zhang}
\author[1]{Jun-Hyuk Kim}
\author[2]{Cho-Jui Hsieh}
\author[1]{Jong-Seok Lee}

\affil[1]{\large\normalfont{School of Integrated Technology, Yonsei University}}
\affil[ ]{\tt\small{\{idearibosome,junhyuk.kim,jong-seok.lee\}@yonsei.ac.kr}}

\affil[2]{\large\normalfont{Department of Computer Science, University of California, Los Angeles}}
\affil[ ]{\tt\small{huanzhang@ucla.edu~~chohsieh@cs.ucla.edu}}

\maketitle
%\thispagestyle{empty}

%%%%%%%%% ABSTRACT
\begin{abstract}
Single-image super-resolution aims to generate a high-resolution version of a low-resolution image, which serves as an essential component in many computer vision applications.
This paper investigates the robustness of deep learning-based super-resolution methods against adversarial attacks, which can significantly deteriorate the super-resolved images without noticeable distortion in the attacked low-resolution images.
It is demonstrated that state-of-the-art deep super-resolution methods are highly vulnerable to adversarial attacks.
Different levels of robustness of different methods are analyzed theoretically and experimentally.
We also present analysis on transferability of attacks, and feasibility of targeted attacks and universal attacks.
\end{abstract}

%%%%%%%%% BODY TEXT
\section{Introduction}

Single-image super-resolution, which is to generate a high-resolution version of a low-resolution image, is one of the popular research areas in recent years.
While simple interpolation methods such as bilinear and bicubic upscaling have been used popularly, the development of deep learning-based approaches, which is triggered by a simple convolutional network model named super-resolution convolutional neural network (SRCNN) \cite{dong2014learning}, offers much better quality of the upscaled images.
The improvement of the super-resolution technique extends its applications to broader areas, including video streaming, surveillance, medical diagnosis, and satellite photography \cite{yue2016image}.

While many deep learning-based super-resolution methods have been introduced, their robustness against intended attacks has not been thoroughly studied.
The vulnerability of deep networks has been an important issue in recent years, since various investigations report that the attack can fool the deep classification models and can cause severe security issues \cite{goodfellow2014explaining,su2018robustness}.
The similar issues can be raised for the super-resolution applications, since the deteriorated outputs can directly affect the reliability and stability of the systems employing super-resolution as their key components.

In this paper, we investigate the robustness of deep learning-based super-resolution against adversarial attacks, which is the first work to the best of our knowledge.
Our attacks generate perturbations in the input images, which are not visually noticeable but can largely deteriorate the quality of the outputs.
The main contributions of this paper can be summarized as follows.
\vspace{-0.3em}
\begin{itemize}
	\setlength\itemsep{-0.1em}
	\item
	We propose three adversarial attack methods for super-resolution, which slightly perturb a given low-resolution image but result in significantly deteriorated output images, including basic, universal, and partial attacks.
	The methods are based on the methods widely used in the image classification tasks, and we optimize them for the super-resolution tasks. 
	\item
	We present thorough analysis of the robustness of the super-resolution methods, by providing experimental results using the adversarial attack methods.
	We employ various state-of-the-art deep learning-based super-resolution methods having variable characteristics in terms of model structure, training objective, and model size.
	\item
	We further investigate the relation of robustness to the model properties and measure the transferability.
	In addition, we provide three advanced topics, including targeted attack, attack-agnostic robustness measurement, and simple defense methods of the attack.
\end{itemize}

%The rest of the paper is organized as follows.
%First, we summarize the related work in Section~\ref{sec:related_work}.
%Then, we explain our attack methods in Section~\ref{sec:attacks}.
%We analyze our experimental results in Section~\ref{sec:experiments}.
%In addition, we discuss advanced topics in Section~\ref{sec:advanced_topics}.
%Finally, we conclude our work in Section~\ref{sec:conclusion}.

%-------------------------------------------------------------------------
\section{Related Work}
\label{sec:related_work}

\noindent \textbf{Super-resolution.}
%Single-image super-resolution traditionally has been investigated by employing feature extraction-based methods such as finding sparse representation \cite{yang2011multitask} and projecting data into subspaces \cite{li2014single}.
Recently, the trend of super-resolution researches has been shifted to the deep learning-based methods.
One of the notable methods that achieve much improved performance is the enhanced deep super-resolution (EDSR) model \cite{lim2017enhanced}.
Later, Zhang \textit{et al.} \cite{zhang2018image} propose a more advanced network model named residual channel attention network (RCAN), which applies an attention mechanism to exploit image features effectively.

While the aforementioned methods focus on achieving high performance in terms of peak signal-to-noise ratio (PSNR), some researchers argue that considering only such a distortion measure does not necessarily enhance perceptual image quality \cite{blau2018perception}.
To deal with this, perceptually optimized super-resolution methods are proposed, which employ generative adversarial networks (GANs) \cite{goodfellow2014generative}.
One of the state-of-the-art methods is the enhanced super-resolution generative adversarial network (ESRGAN) \cite{wang2018esrgan}, which generates more visually appealing outputs than other conventional methods, even though the PSNR values are lower.
Choi \textit{et al.} \cite{choi2018deep} develop the four-pass perceptual enhanced upscaling super-resolution (4PP-EUSR) method, which considers both the quantitative and perceptual quality to obtain more natural upscaled images.

Since super-resolution is also a useful component in mobile applications, some studies focus on economizing the computational resource while reasonable performance is maintained.
For instance, Ahn \textit{et al.} \cite{ahn2018fast} propose the cascading residual network (CARN) and its mobile version (CARN-M), which employ cascading residual blocks with shared model parameters.
\\[-0.8\baselineskip]

\noindent \textbf{Adversarial attack.}
Recent studies show that deep image classifiers are vulnerable to various adversarial attacks.
Szegedy \textit{et al.} \cite{szegedy2013intriguing} propose an optimization-based attack method that aims to minimize the amount of perturbation with changing the classification result of a classifier.
Goodfellow \textit{et al.} \cite{goodfellow2014explaining} develop the fast gradient sign method (FGSM), which uses the sign of the gradients that are obtained from the classifier.
Kurakin \textit{et al.} \cite{kurakin2016adversarial} extend it to an iterative approach (I-FGSM), which shows higher success rate of the attack than FGSM.
These attacks are known as strong attack methods that can fool almost every state-of-the-art image classifier with high success rate \cite{su2018robustness}.

Some studies provide in-depth analysis of the robustness of deep learning models.
Liu \textit{et al.} \cite{liu2016delving} measure transferability of the adversarial images, which is to find out whether the perturbations found for a classifier also work for another classifier.
Moosavi-Dezfooli \textit{et al.} \cite{moosavi2017universal} investigate a universal perturbation that can be applied to all images in a given dataset.
Weng \textit{et al.} \cite{weng2018evaluating} proposed a theoretical robustness measure, which does not depend on a specific attack method.
\\[-0.8\baselineskip]

\noindent \textbf{Adversarial attack on super-resolution.}
Recently, combining the super-resolution tasks with adversarial attacks has emerged.
Mustafa \textit{et al.} \cite{mustafa2019image} presents a method employing super-resolution to defense deep image classifiers against adversarial attacks.
Yin \textit{et al.} \cite{yin2018deep} employ the adversarial attack on super-resolution to fool the subsequent computer vision tasks.
However, these studies investigate the effectiveness of the adversarial attack for other tasks rather than the super-resolution task itself, including image classification, style transfer, and image captioning, where the super-resolution is used as a pre-processing step before the main tasks.
Therefore, the robustness of super-resolution itself against adversarial attacks, which is investigated in this paper, has not been addressed previously.

%-------------------------------------------------------------------------
\section{Attacks on Super-Resolution}
\label{sec:attacks}

\subsection{Basic attack}
\label{sec:basic_attack}

The goal of the adversarial attack on super-resolution models is to inject a small amount of perturbation in the given input image so that the perturbation is not visually perceivable but results in significant deterioration in the super-resolved output.
To do this, we develop an algorithm based on the idea of I-FGSM \cite{kurakin2016adversarial}, which is one of the most widely used strong attacks for classification models.

Let $\mathbf{X}_{0}$ denote the original low-resolution input image and $\mathbf{X}$ denote the attacked version of $\mathbf{X}_{0}$.
From these images, we obtain the super-resolved high-resolution images ${f}(\mathbf{X}_{0})$ and ${f}(\mathbf{X})$, respectively, via a given super-resolution model ${f}(\cdot)$.
Our objective is to maximize the amount of deterioration in the super-resolved output, which can be defined as:
\begin{equation}
\label{eq:basic_attack_sr_loss}
\mathcal{L} ( \mathbf{X}, \mathbf{X}_{0} ) = || {f}(\mathbf{X}) - {f}(\mathbf{X}_{0}) ||_{2}.
\end{equation}
To find an $\mathbf{X}$ to minimize \eqref{eq:basic_attack_sr_loss} with bounded $\ell_\infty$-norm constraint ($\|\mathbf{X}-\mathbf{X}_0\|_\infty \leq \alpha$),  
we adopt the I-FGSM update rule that iteratively updates $\mathbf{X}$ by: 
%to find the adversarial example iteratively as:
\begin{equation}
\label{eq:basic_attack_iteration_tilde}
\widetilde{\mathbf{X}}_{n+1} = \mathrm{clip}_{0, 1} \Big( \mathbf{X}_{n} + \frac{\alpha}{T}~\mathrm{sgn} \big( \nabla \mathcal{L} ( \mathbf{X}_{n}, \mathbf{X}_{0} ) \big) \Big)
\end{equation}
\begin{equation}
\label{eq:basic_attack_iteration}
\mathbf{X}_{n+1} = \mathrm{clip}_{-\alpha, \alpha} ( \widetilde{\mathbf{X}}_{n+1} - \mathbf{X}_{0} ) + \mathbf{X}_{0}
\end{equation}
where $T$ is the number of iterations, $\mathrm{sgn} \big( \nabla \mathcal{L} ( \mathbf{X}_{n}, \mathbf{X}_{0} ) \big)$ is the sign of the gradient of (\ref{eq:basic_attack_sr_loss}), and
\begin{equation}
\mathrm{clip}_{a, b}(\mathbf{X}) = \mathrm{min} \big( \mathrm{max} ( \mathbf{X}, a ), b \big).
\end{equation}
The term $\alpha$ not only controls the amount of contribution that the calculated gradient provides at each iteration, but also limits the maximum amount of perturbation to prevent noticeable changes of the attacked input image.
The final adversarial example is obtained by $\mathbf{X} = \mathbf{X}_{T}$.

\subsection{Universal attack}

Although an adversarial image can be found for each image as in Section~\ref{sec:basic_attack}, it is also possible to find an \textit{image-agnostic} adversarial perturbation, which can affect any input image for a certain super-resolution method \cite{moosavi2017universal}.
We apply this concept in our study by altering the formulation of our basic attack as follows.

Assume that there are $K$ images in the dataset, where the $k$-th image is denoted as $\mathbf{X}_{0}^{k}$.
With a universal perturbation $\Delta$, we can obtain the adversarial images as:
\begin{equation}
\mathbf{X}^{k} = \mathrm{clip}_{0, 1} ( \mathbf{X}_{0}^{k} + \Delta ).
\end{equation}
Then, we compute the average amount of deterioration as:
\begin{equation}
\mathcal{F}(\Delta) = \frac{1}{K} \sum_{k=1}^{K} \mathcal{L} ( \mathbf{X}^{k}, \mathbf{X}_{0}^{k} ).
\end{equation}
Starting from ${\Delta}_{0} = 0$, the universal perturbation is updated iteratively as:
\begin{equation}
{\Delta}_{n+1} = \mathrm{clip}_{-\alpha, \alpha} \Big( {\Delta}_{n} + \frac{\alpha}{T}~\mathrm{sgn} \big( \nabla \mathcal{F} ( {\Delta}_{n} ) \big) \Big).
\end{equation}
The final universal perturbation is obtained by $\Delta = {\Delta}_{T}$.

\begin{table}[]
	\small
	\setlength{\tabcolsep}{0.5em}
	\renewcommand{\arraystretch}{1.05}
	\begin{center}
		\begin{tabular}{l|rrr}
			\textbf{Method} & \textbf{\# parameters} & \textbf{\# layers} & \textbf{GAN-based} \\
			\hline
			EDSR \cite{lim2017enhanced} & 43.1M & 69 & - \\
			EDSR-baseline \cite{lim2017enhanced} & 1.5M & 37 & - \\
			RCAN \cite{zhang2018image} & 15.6M & 815 & - \\
			4PP-EUSR \cite{choi2018deep} & 6.3M & 95 & \checkmark \\
			ESRGAN \cite{wang2018esrgan} & 16.7M & 351 & \checkmark \\
			RRDB \cite{wang2018esrgan} & 16.7M & 351 & - \\
			CARN \cite{ahn2018fast} & 1.1M & 34 & - \\
			CARN-M \cite{ahn2018fast} & 0.3M & 43 & -
		\end{tabular}
	\end{center}
	\caption{Properties of the super-resolution methods.}
	\label{table:sr_methods_properties}
\end{table}

\begin{figure*}[t]
	\begin{center}
		\centering
		\begin{minipage}[b]{0.33\linewidth}
			\centering
			\centerline{\includegraphics[width=1.0\linewidth]{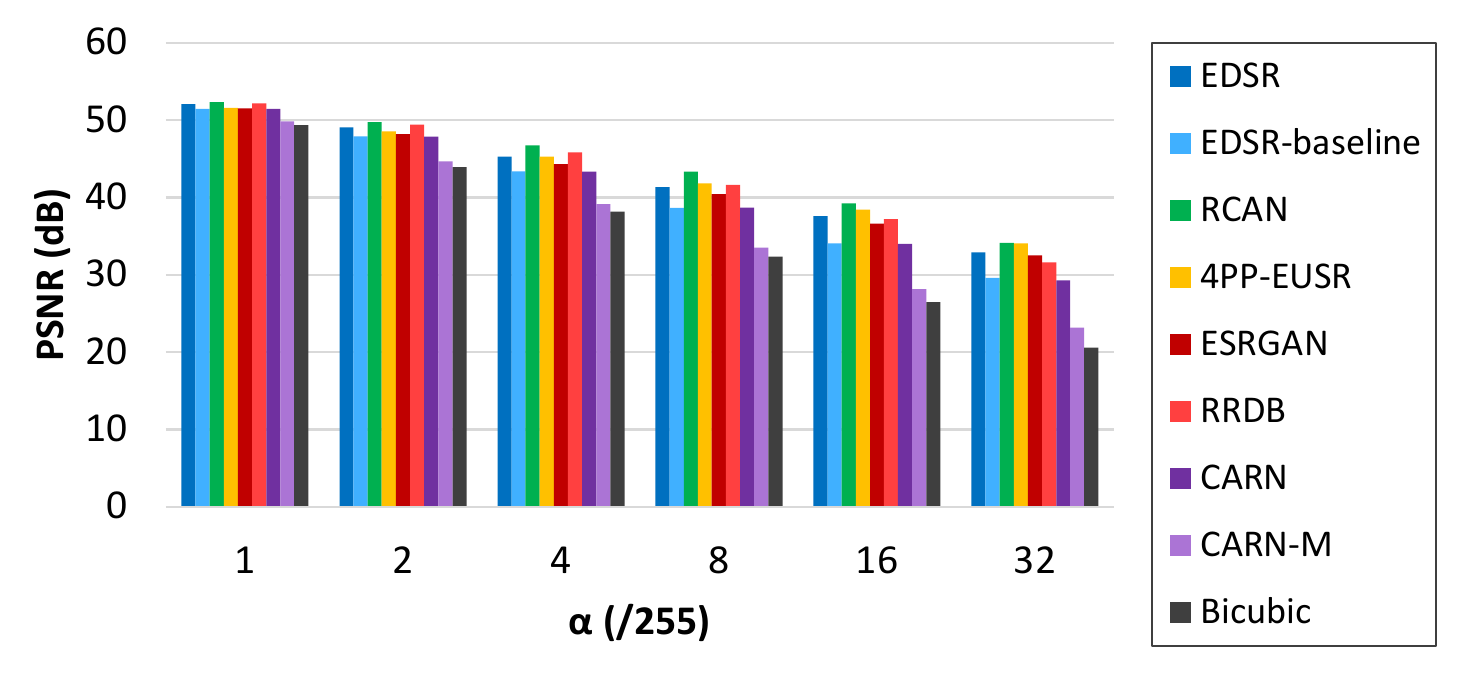}}
			\centerline{(a) Set5, LR}
		\end{minipage}
		\begin{minipage}[b]{0.33\linewidth}
			\centering
			\centerline{\includegraphics[width=1.0\linewidth]{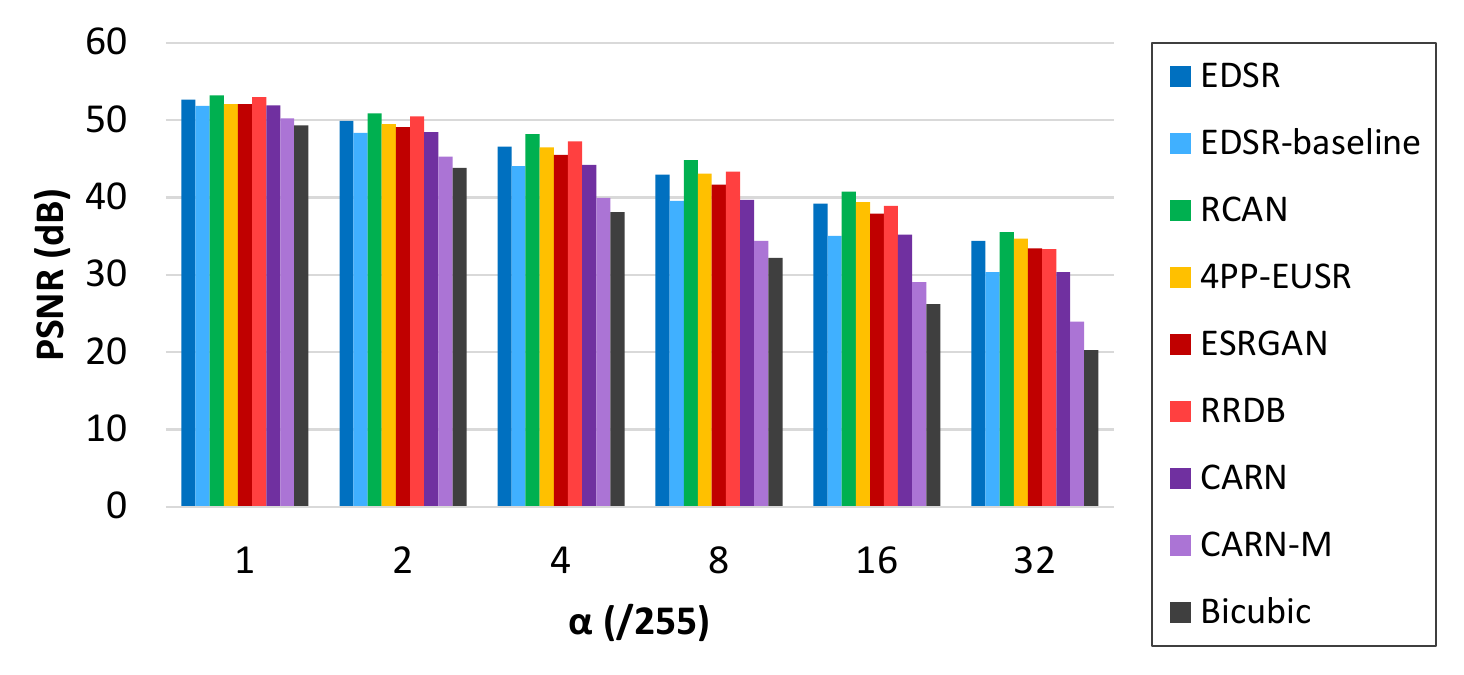}}
			\centerline{(b) Set14, LR}
		\end{minipage}
		\begin{minipage}[b]{0.33\linewidth}
			\centering
			\centerline{\includegraphics[width=1.0\linewidth]{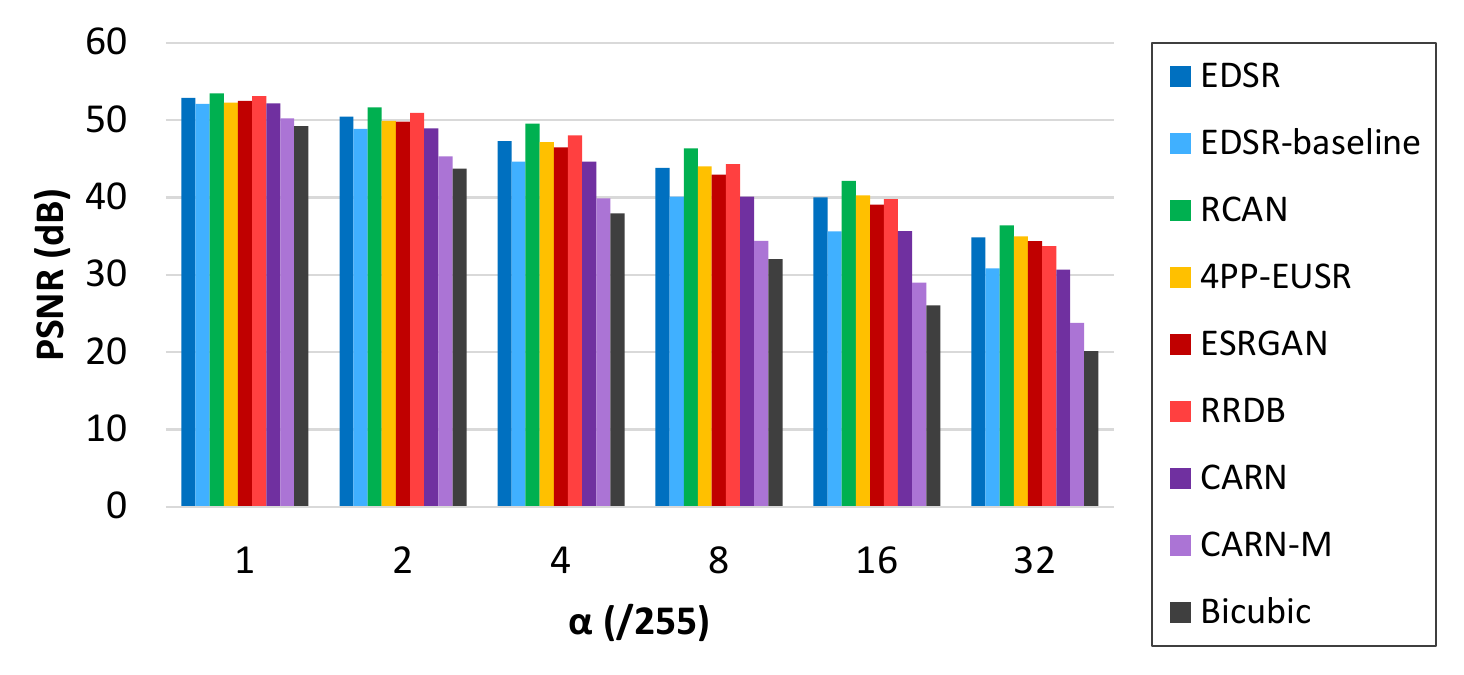}}
			\centerline{(c) BSD100, LR}
		\end{minipage}
		\begin{minipage}[b]{0.33\linewidth}
			\centering
			\centerline{\includegraphics[width=1.0\linewidth]{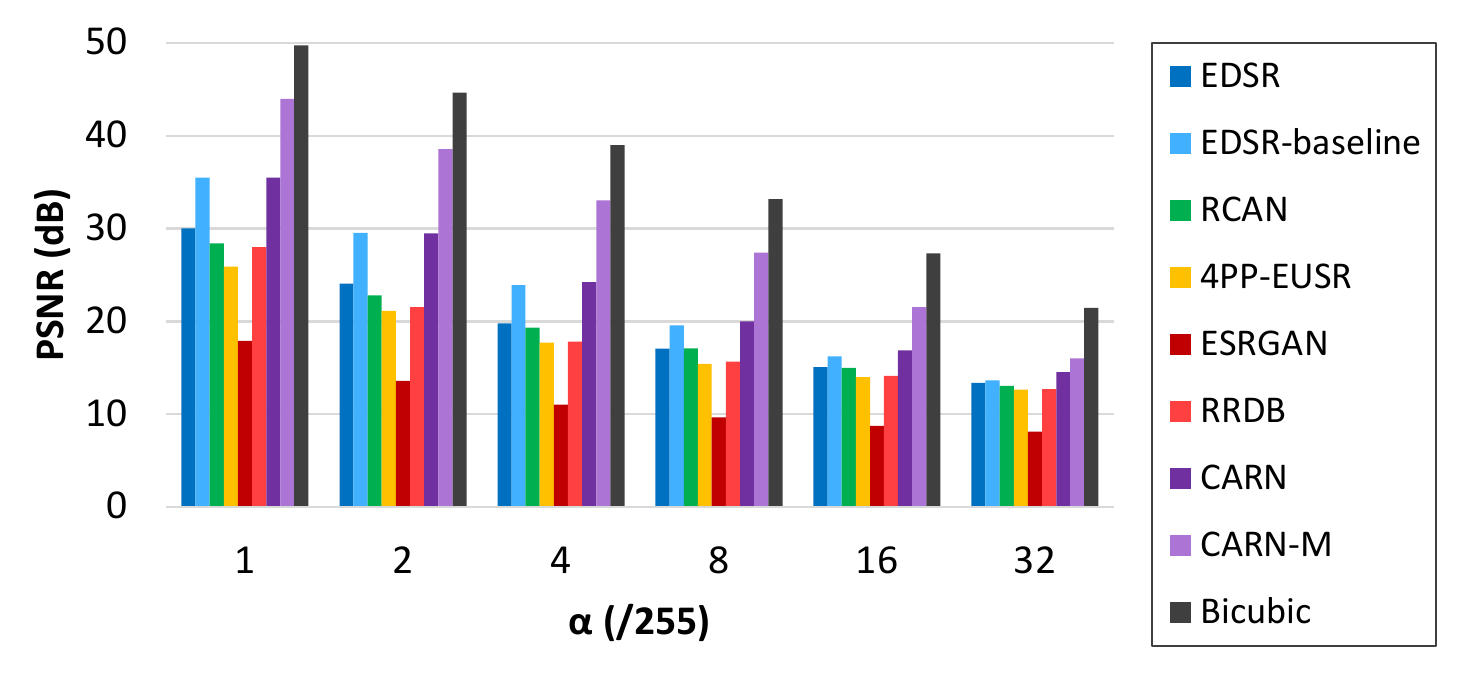}}
			\centerline{(d) Set5, SR}
		\end{minipage}
		\begin{minipage}[b]{0.33\linewidth}
			\centering
			\centerline{\includegraphics[width=1.0\linewidth]{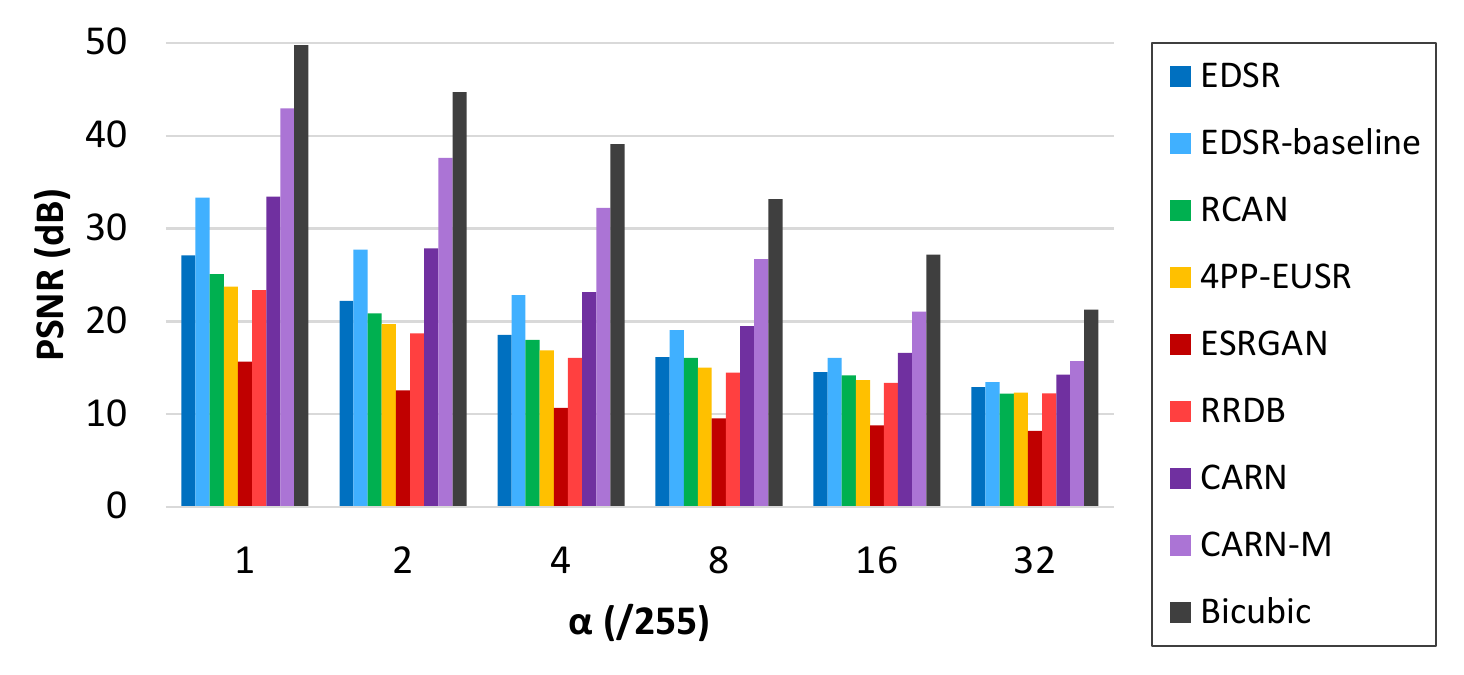}}
			\centerline{(e) Set14, SR}
		\end{minipage}
		\begin{minipage}[b]{0.33\linewidth}
			\centering
			\centerline{\includegraphics[width=1.0\linewidth]{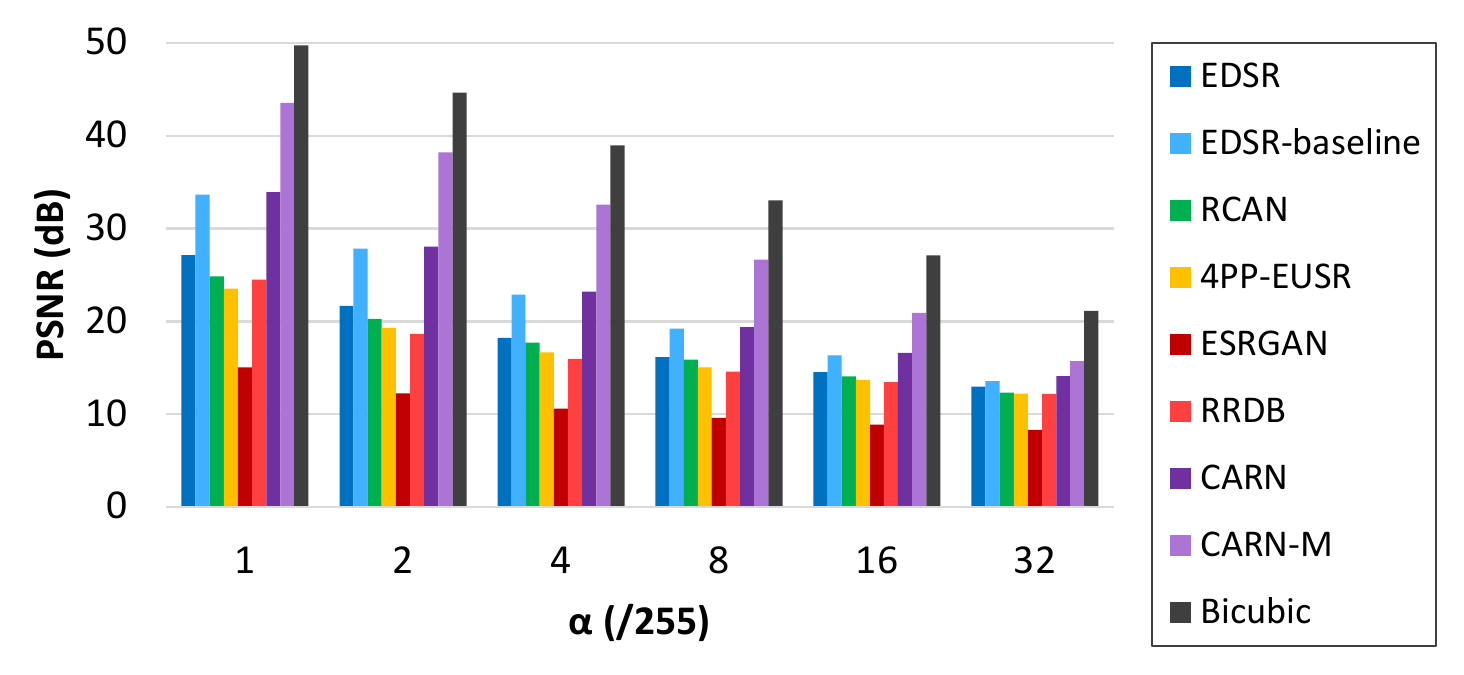}}
			\centerline{(f) BSD100, SR}
		\end{minipage}
	\end{center}
	\caption{Comparison of the PSNR values of low-resolution (LR) and super-resolved (SR) images with respect to different $\alpha$ values for the basic attack on the Set5 \cite{bevilacqua2012low}, Set14 \cite{zeyde2010single}, and BSD100 \cite{martin2001database} datasets.}
	\label{fig:basic_alpha_psnr}
\end{figure*}

\subsection{Partial attack}

The basic attack in Section~\ref{sec:basic_attack} finds a perturbation covering the whole region of the given image.
We further investigate the robustness of the super-resolution methods by attacking only some part of the image, but measuring the amount of deterioration in the region that is not being attacked.
With this experiment, we can examine how much the perturbation \textit{permeates} into the adjacent regions during super-resolution.

Let $\mathbf{M}$ denote a binary mask of the perturbation $\Delta$, where only the region to be attacked is set to 1.
The masked perturbation is $\Delta \circ \mathbf{M}$, where $\circ$ denotes the element-wise multiplication.
Then, (\ref{eq:basic_attack_iteration_tilde}) can be modified as:
\begin{equation}
\label{eq:partial_attack_iteration_tilde}
\widetilde{\mathbf{X}}_{n+1} = \mathrm{clip}_{0, 1} \Big( \mathbf{X}_{n} + \frac{\alpha}{T}~\mathrm{sgn} \big( \nabla \mathcal{L}_{\mathbf{M}} ( \mathbf{X}_{n}, \mathbf{X}_{0} ) \big) \circ \mathbf{M} \Big)
\end{equation}
where
\begin{equation}
\label{eq:partial_attack_sr_loss}
\mathcal{L}_{\mathbf{M}} ( \mathbf{X}, \mathbf{X}_{0} ) = {\big|\big| \big( {f}(\mathbf{X}) - {f}(\mathbf{X}_{0}) \big) \circ (1 - \mathbf{M}_{H}) \big|\big|}_{2}.
\end{equation}
In (\ref{eq:partial_attack_sr_loss}), $\mathbf{M}_{H}$ is a high-resolution counterpart of $\mathbf{M}$.
The term $(1-\mathbf{M}_{H})$ ensures that the amount of deterioration is calculated only on the unperturbed regions.
The final adversarial example is obtained by $\mathbf{X} = \mathbf{X}_{T}$.

\begin{figure*}[t]
	\begin{center}
		\centering
		\begin{minipage}[b]{0.19\linewidth}
			\centering
			\centerline{\includegraphics[width=0.98\linewidth]{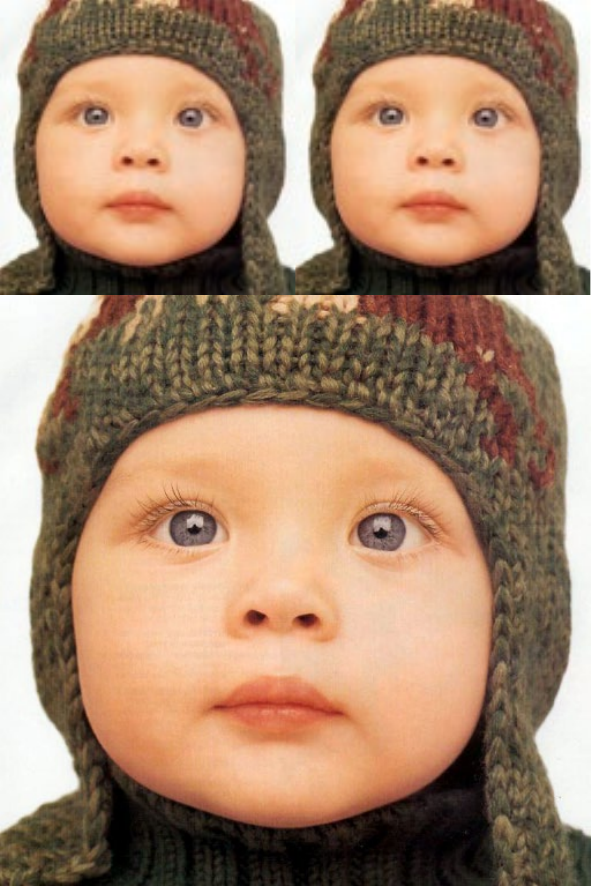}}
			\centerline{Ground-truth}
		\end{minipage}
		\begin{minipage}[b]{0.19\linewidth}
			\centering
			\centerline{\includegraphics[width=0.98\linewidth]{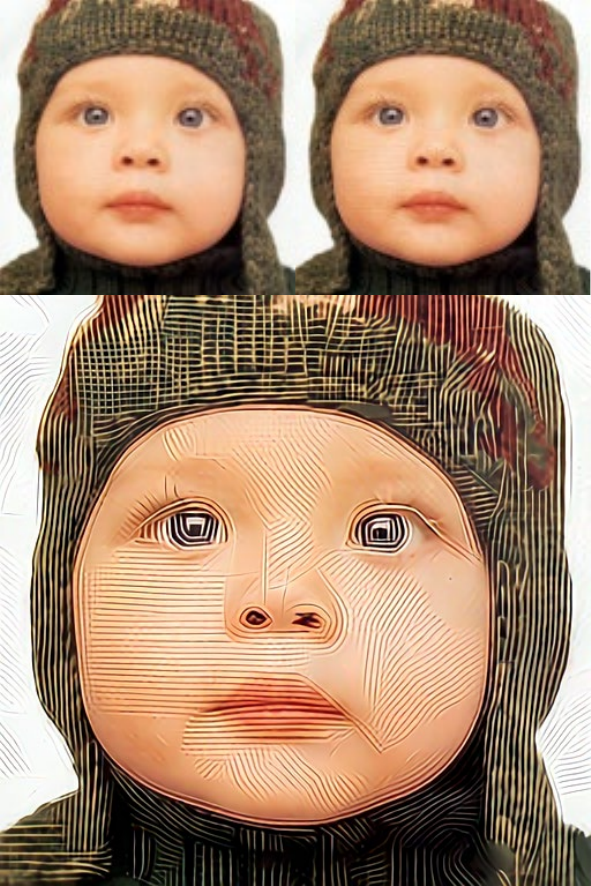}}
			\centerline{EDSR}
		\end{minipage}
		\begin{minipage}[b]{0.19\linewidth}
			\centering
			\centerline{\includegraphics[width=0.98\linewidth]{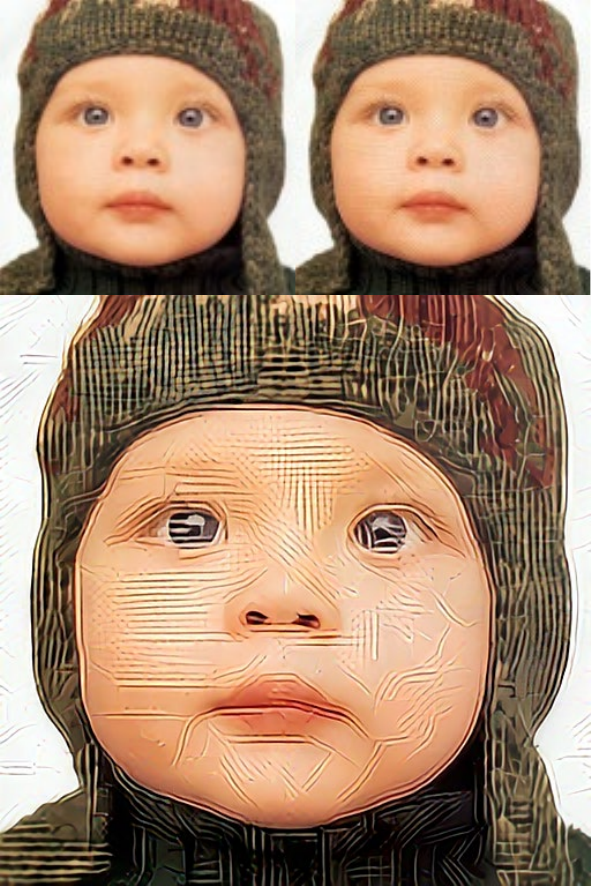}}
			\centerline{EDSR-baseline}
		\end{minipage}
		\begin{minipage}[b]{0.19\linewidth}
			\centering
			\centerline{\includegraphics[width=0.98\linewidth]{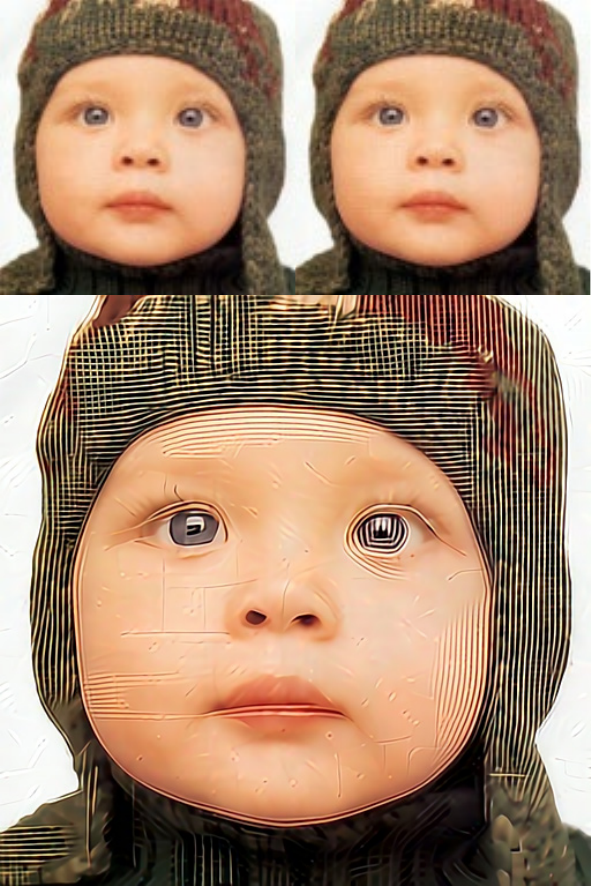}}
			\centerline{RCAN}
		\end{minipage}
		\begin{minipage}[b]{0.19\linewidth}
			\centering
			\centerline{\includegraphics[width=0.98\linewidth]{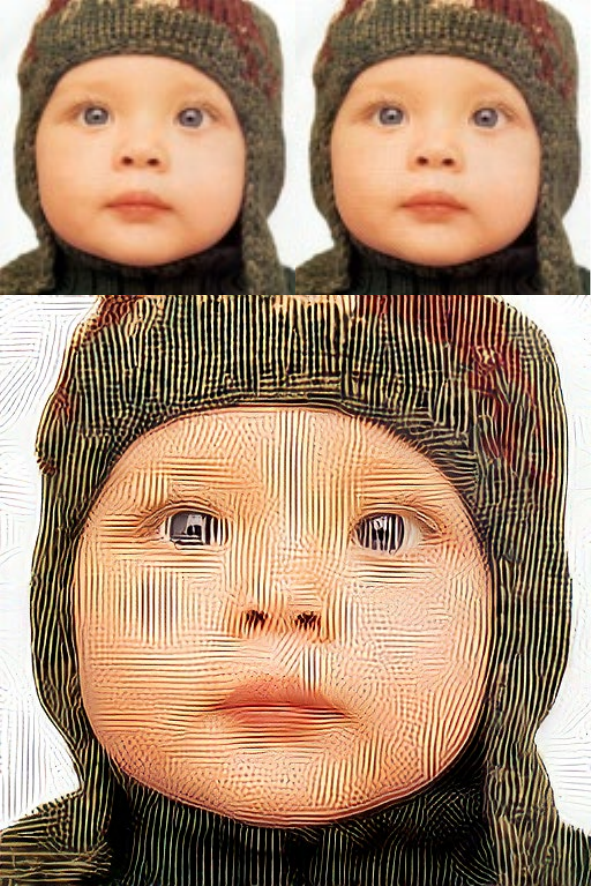}}
			\centerline{4PP-EUSR}
		\end{minipage}
		\\ \medskip
		\begin{minipage}[b]{0.19\linewidth}
			\centering
			\centerline{\includegraphics[width=0.98\linewidth]{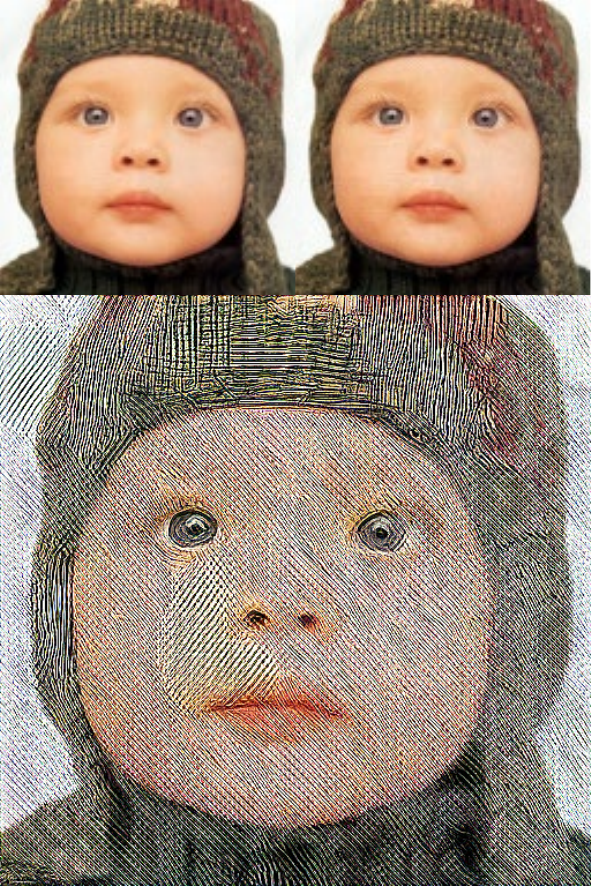}}
			\centerline{ESRGAN}
		\end{minipage}
		\begin{minipage}[b]{0.19\linewidth}
			\centering
			\centerline{\includegraphics[width=0.98\linewidth]{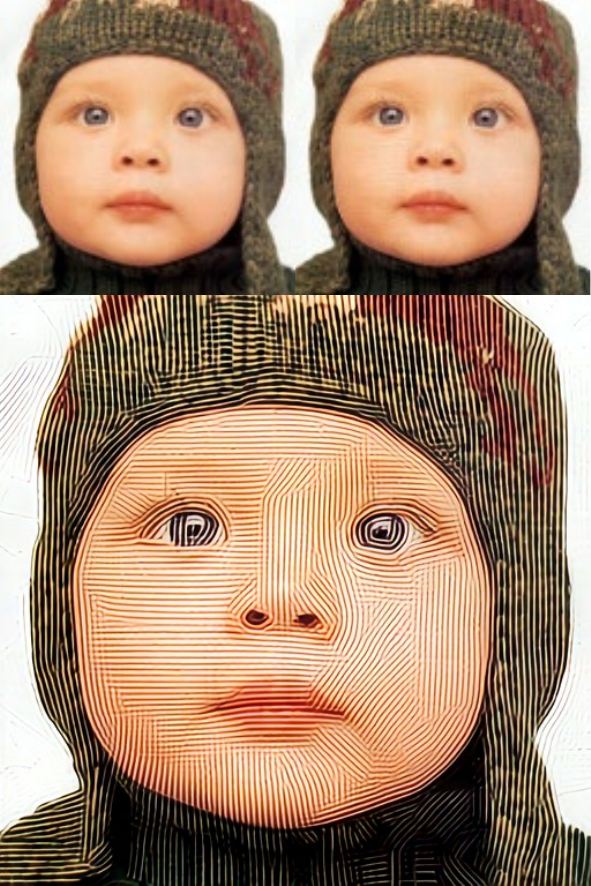}}
			\centerline{RRDB}
		\end{minipage}
		\begin{minipage}[b]{0.19\linewidth}
			\centering
			\centerline{\includegraphics[width=0.98\linewidth]{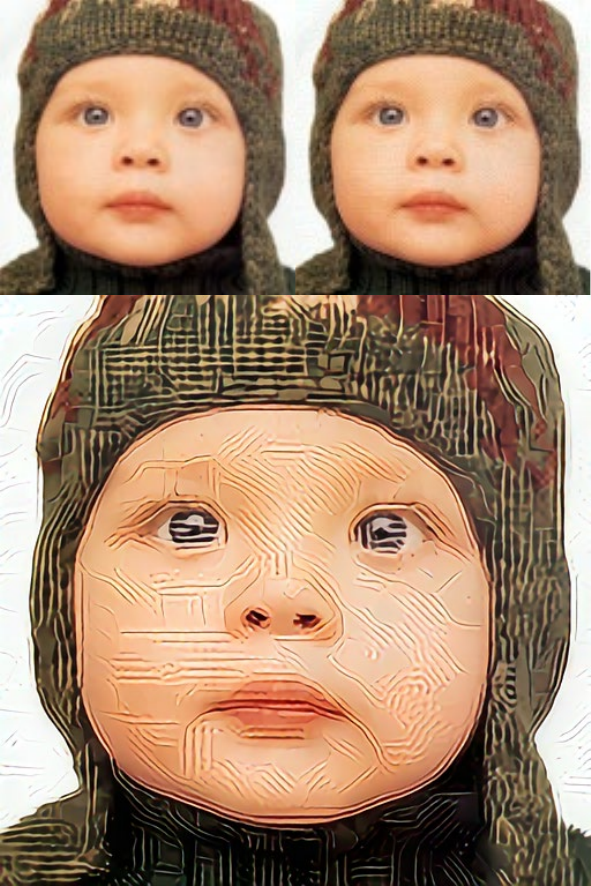}}
			\centerline{CARN}
		\end{minipage}
		\begin{minipage}[b]{0.19\linewidth}
			\centering
			\centerline{\includegraphics[width=0.98\linewidth]{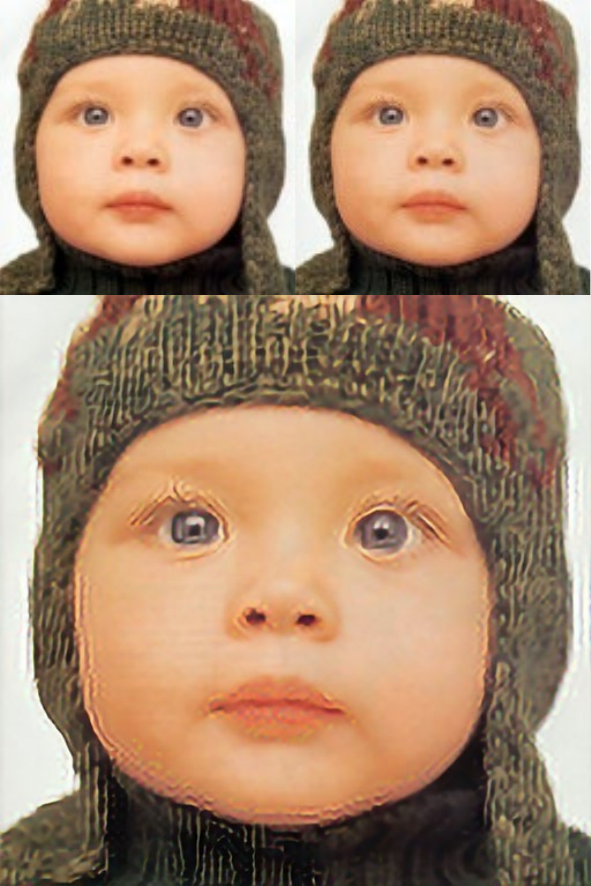}}
			\centerline{CARN-M}
		\end{minipage}
		\begin{minipage}[b]{0.19\linewidth}
			\centering
			\centerline{\includegraphics[width=0.98\linewidth]{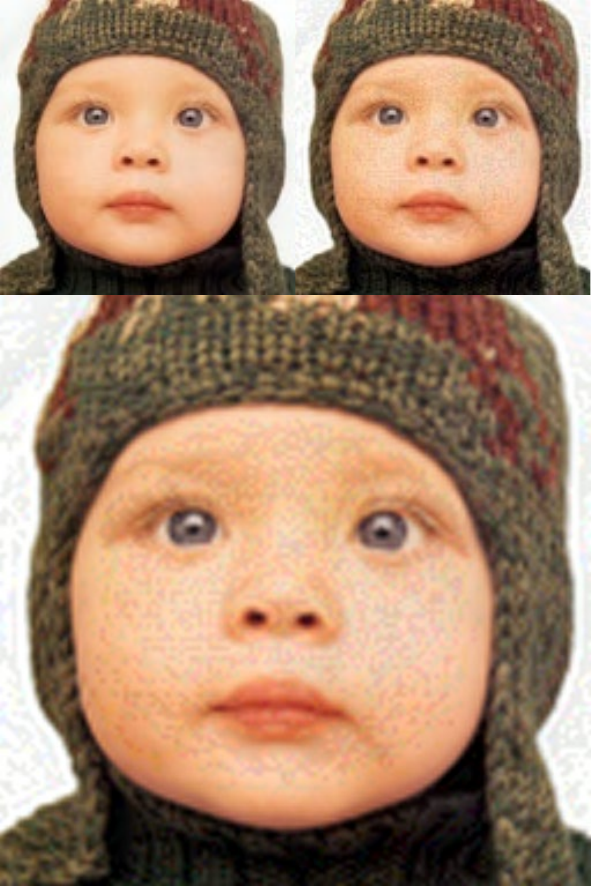}}
			\centerline{Bicubic}
		\end{minipage}
	\end{center}
	\caption{Visual comparison of the super-resolved outputs for the inputs attacked with $\alpha=8/255$. In each case, (top-left) is the original input in Set5 \cite{bevilacqua2012low}, (top-right) is the adversarial input, and (bottom) is the output obtained from the adversarial input. The input images are enlarged two times for better visualization.}
	\label{fig:basic_example}
\end{figure*}

%-------------------------------------------------------------------------
\section{Experimental Results}
\label{sec:experiments}

\noindent \textbf{Datasets.}
We employ three image datasets that are widely used for benchmarking super-resolution methods: Set5 \cite{bevilacqua2012low}, Set14 \cite{zeyde2010single}, and BSD100 \cite{martin2001database}. %\footnote{Due to the space limit, the results for some datasets are omitted for some cases, which can be found in the Supplementary Material.}
Each dataset consists of 5, 14, and 100 images, respectively.
\\[-0.8\baselineskip]

\noindent \textbf{Super-resolution methods.}
We consider eight state-of-the-art deep learning-based super-resolution methods having various model sizes and properties, including EDSR \cite{lim2017enhanced}, EDSR-baseline \cite{lim2017enhanced}, RCAN \cite{zhang2018image}, 4PP-EUSR \cite{choi2018deep}, ESRGAN \cite{wang2018esrgan}, RRDB \cite{wang2018esrgan}, CARN \cite{ahn2018fast}, and CARN-M \cite{ahn2018fast}.
Table~\ref{table:sr_methods_properties} shows their characteristics in terms of the number of model parameters, the number of convolutional layers, and whether to employ GANs for training.
EDSR-baseline is a smaller version of EDSR, RRDB is an alternative version of ESRGAN trained without the GAN, and CARN-M is a lightweight version of CARN in terms of the number of model parameters.
In addition, we also consider the bicubic interpolation to compare its robustness against the adversarial attacks with that of the deep learning-based methods.
We consider a scaling factor of 4 for all the super-resolution methods.
In addition, we employ the pre-trained models provided by the original authors.
\\[-0.8\baselineskip]

\begin{figure*}[t]
	\begin{center}
		\centering
		\begin{minipage}[b]{0.45\linewidth}
			\centering
			\centerline{\includegraphics[width=0.88\linewidth]{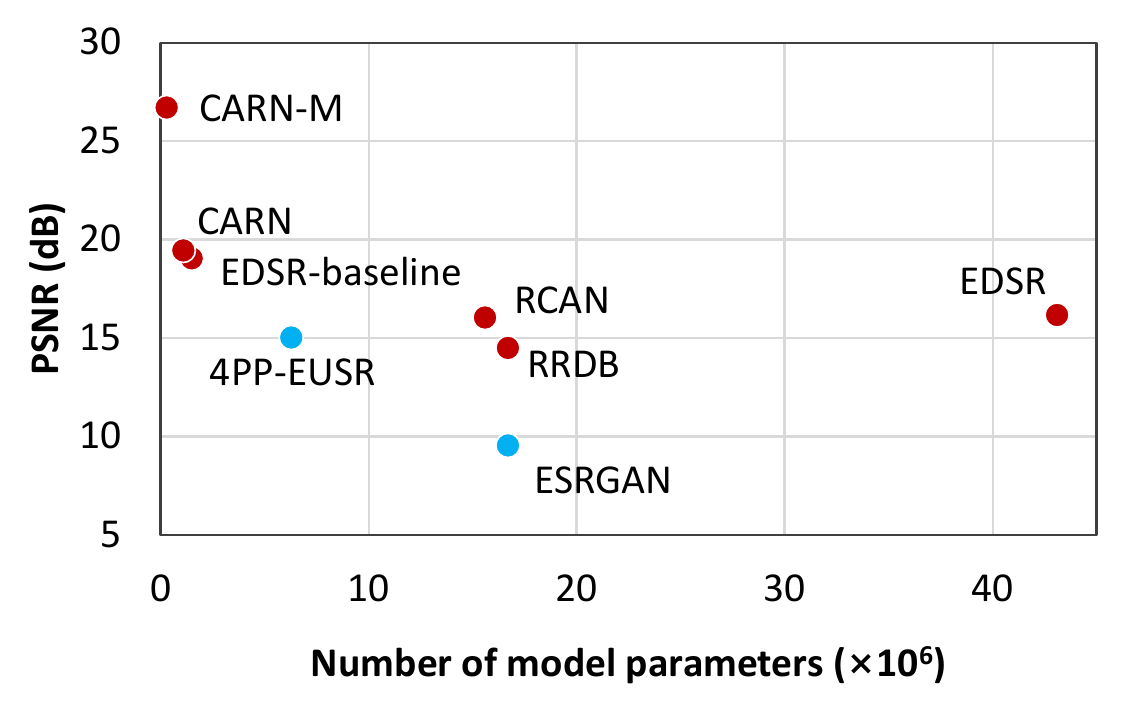}}
			\centerline{(a)}
		\end{minipage}
		\begin{minipage}[b]{0.45\linewidth}
			\centering
			\centerline{\includegraphics[width=0.88\linewidth]{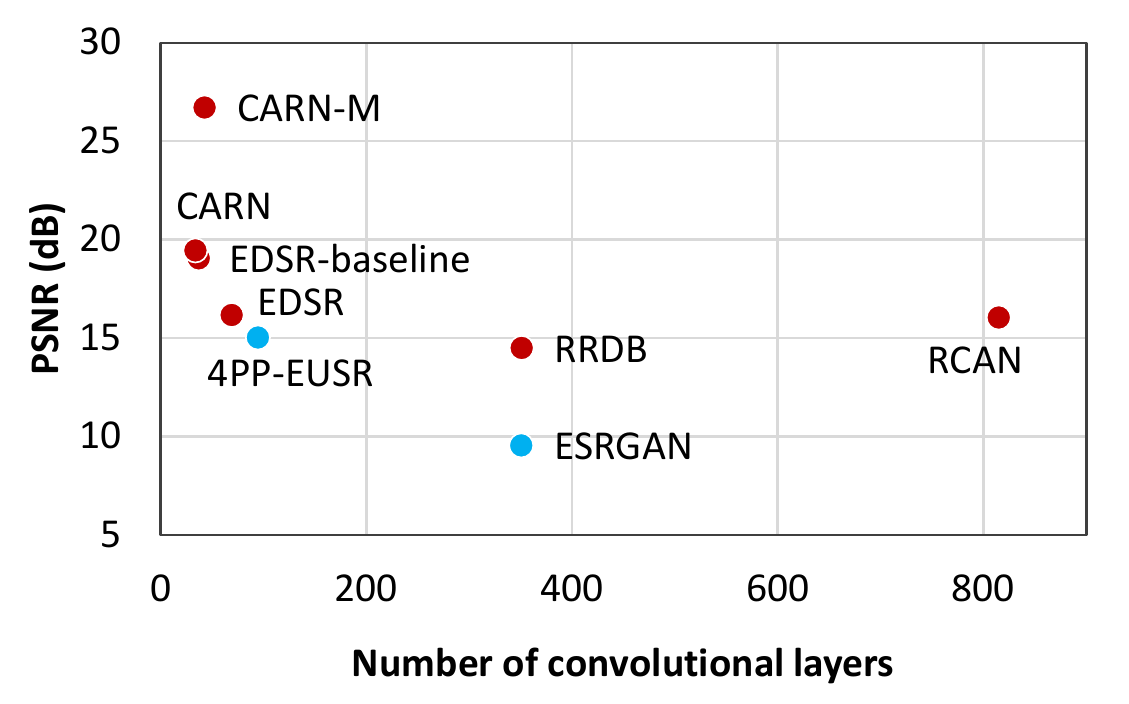}}
			\centerline{(b)}
		\end{minipage}
	\end{center}
	\caption{Comparison of the PSNR values of SR images for BSD100 \cite{martin2001database} with respect to the model sizes in terms of (a) the number of model parameters and (b) the number of convolutional layers ($\alpha=8/255$). Blue and red colors indicate the models trained with and without GANs, respectively.}
	\label{fig:relation_model_size}
\end{figure*}

\begin{figure}[t]
	\begin{center}
		\centering
		\includegraphics[width=1.0\linewidth]{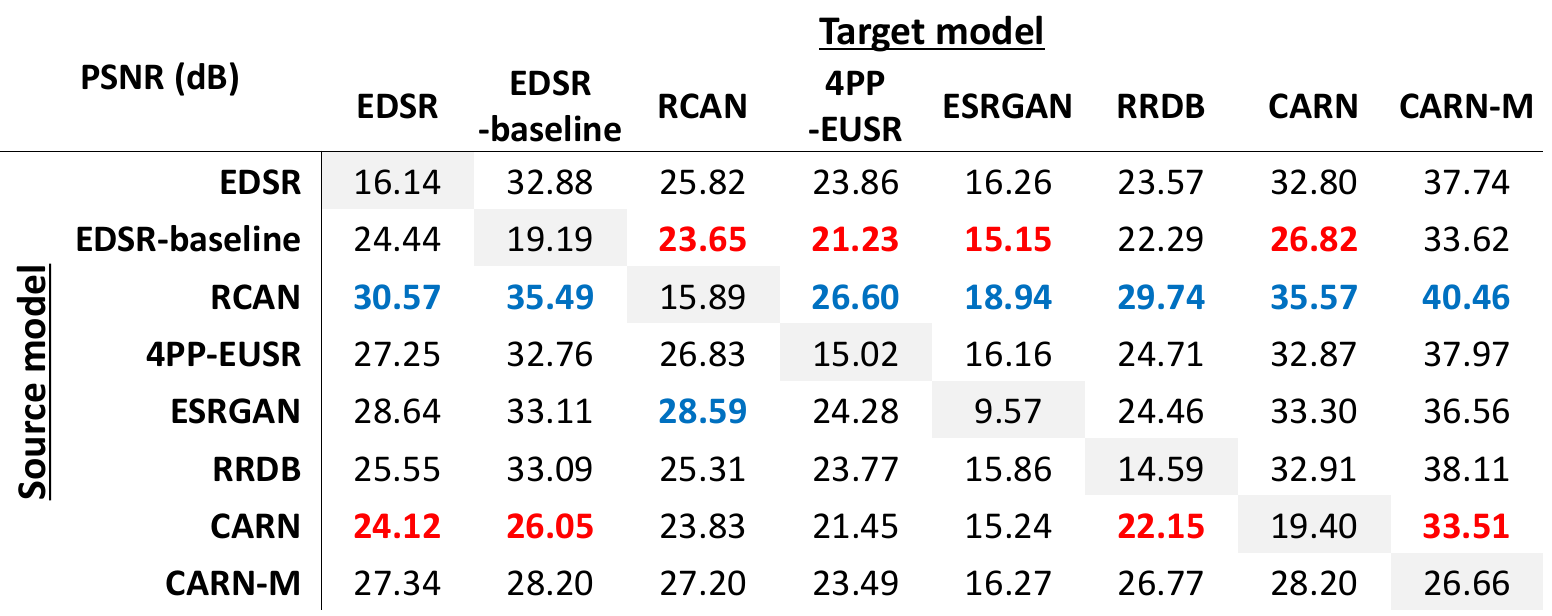}
	\end{center}
	\caption{Comparison of the transferability in terms of PSNR for the BSD100 dataset \cite{martin2001database} when $\alpha=8/255$. Red and blue colors indicate the lowest and highest PSNR values (except the diagonal cells) for each target model, respectively.}
	\label{fig:basic_transferability_psnr}
\end{figure}

\noindent \textbf{Implementation details.}
Our adversarial attack methods are implemented on the TensorFlow framework \cite{abadi2016tensorflow}.
For all the attack methods, we set $\alpha \in \{1/255, 2/255, 4/255, 8/255, 16/255, 32/255\}$ and $T=50$.
For the universal attack, a perturbation $\Delta$ with a fixed spatial resolution is required in order to apply it to all images in a dataset.
Therefore, we crop the center region of each input image with a fixed resolution.
For the partial attack, we set the mask $\mathbf{M}$ so as to attack the central part of the input image, i.e.,
\begin{equation}
\mathbf{M}_{(x, y)} = 
\begin{cases}
1 & \textrm{if } \frac{w}{4} \le x < \frac{3w}{4}, \frac{h}{4} \le y < \frac{3h}{4} \\
0 & \textrm{otherwise}
\end{cases}
\end{equation}
where $\mathbf{M}_{(x, y)}$ is the value of $\mathbf{M}$ at $(x, y)$, and $w$ and $h$ are the width and height of the input image, respectively.
\\[-0.8\baselineskip]

\noindent \textbf{Performance measurement.}
We measure the robustness of the super-resolution methods against our adversarial attack methods in terms of PSNR.
For low-resolution (LR) images, we calculate the PSNR values between the original and attacked images, i.e., $\mathbf{X}_{0}$ and $\mathbf{X}$.
For super-resolved (SR) images, PSNR is measured between the output images obtained from the original and attacked input images, i.e., $f(\mathbf{X}_{0})$ and $f(\mathbf{X})$.
We report the averaged PSNR values for each dataset.
For the partial attack, we calculate the PSNR values only for the outer region of the output image that corresponds to the masked region during the attack.

\subsection{Basic attack}
\label{sec:basic_attack_result}

\figurename~\ref{fig:basic_alpha_psnr} compares the performance of the super-resolution methods in terms of PSNR for the I-FGSM attack explained in Section~\ref{sec:basic_attack}.
As $\alpha$ increases, quality degradation becomes severe in both the LR and SR images.
However, it is much more significant in the SR images than the LR images (i.e., lower PSNR values) except for the bicubic interpolation.
For example, on the Set5 dataset, the PSNR values of LR and SR images for the EDSR model are 41.37 and 17.05 dB, respectively, when $\alpha=8/255$.
Note that two images having a PSNR value higher than 30 dB can be regarded as visually identical images \cite{huang2011robust}.

\figurename~\ref{fig:basic_example} shows example LR and SR images for $\alpha=8/255$.
Overall, there is no obvious difference between the original and perturbed input images for all the super-resolution methods.
However, significant quality deterioration can be observed in the SR images for all methods.
ESRGAN shows the worst visual quality with degradation in all parts of the SR image, which can also be observed as the lowest PSNR values in Figures~\ref{fig:basic_alpha_psnr}d, \ref{fig:basic_alpha_psnr}e, and \ref{fig:basic_alpha_psnr}f.
For the other super-resolution models, fingerprint-like patterns are observed.
This proves that all the deep learning-based super-resolution methods are highly vulnerable against the adversarial attack.
In comparison, the bicubic method, although having lower super-resolution quality on clean data, is much more robust compared with the deep learning-based approaches.
\\[-0.8\baselineskip]

\begin{figure*}[t]
	\begin{center}
		\centering
		\begin{minipage}[b]{0.45\linewidth}
			\centering
			\centerline{\includegraphics[width=0.92\linewidth]{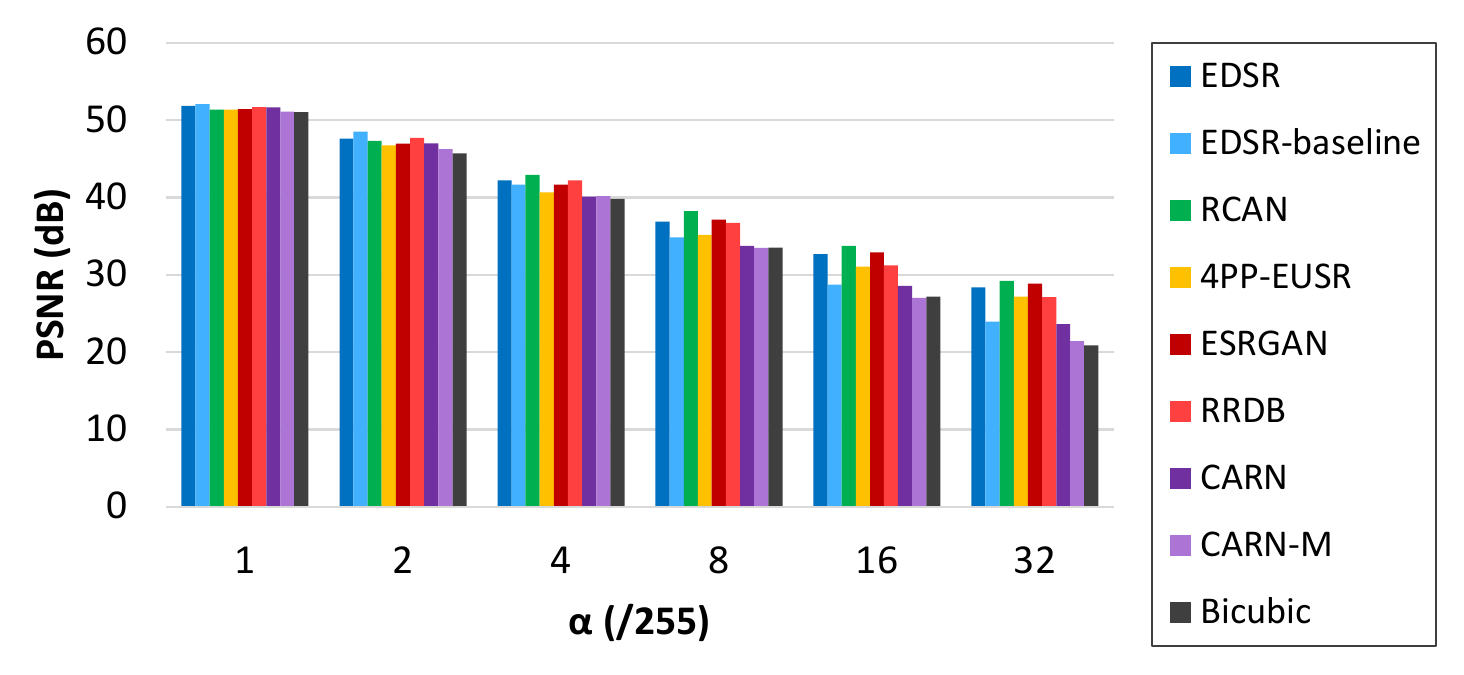}}
			\centerline{(a) LR}
		\end{minipage}
		\begin{minipage}[b]{0.45\linewidth}
			\centering
			\centerline{\includegraphics[width=0.92\linewidth]{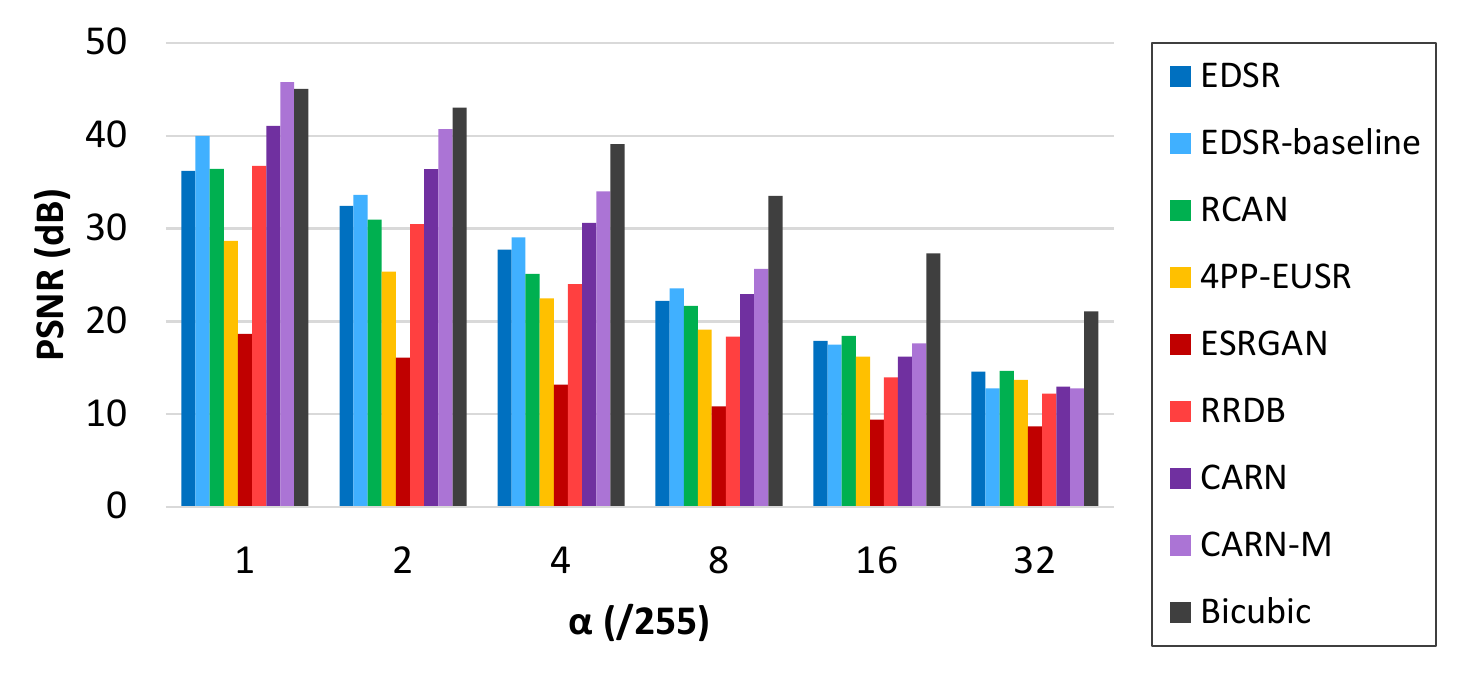}}
			\centerline{(b) SR}
		\end{minipage}
	\end{center}
	\caption{Comparison of the PSNR values of LR and SR images with respect to different $\alpha$ values for the universal attack on the BSD100 dataset \cite{martin2001database}.}
	\label{fig:universal_alpha_psnr}
\end{figure*}

\begin{figure*}[t]
	\begin{center}
		\centering
		\begin{minipage}[b]{0.16\linewidth}
			\centering
			\centerline{\includegraphics[width=1.0\linewidth]{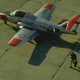}}
			\centerline{(a)}
		\end{minipage}
		\begin{minipage}[b]{0.16\linewidth}
			\centering
			\centerline{\includegraphics[width=1.0\linewidth]{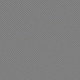}}
			\centerline{(b)}
		\end{minipage}
		\begin{minipage}[b]{0.16\linewidth}
			\centering
			\centerline{\includegraphics[width=1.0\linewidth]{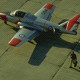}}
			\centerline{(c)}
		\end{minipage}
		\begin{minipage}[b]{0.16\linewidth}
			\centering
			\centerline{\includegraphics[width=1.0\linewidth]{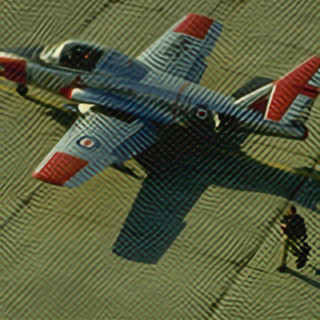}}
			\centerline{(d)}
		\end{minipage}
		\begin{minipage}[b]{0.16\linewidth}
			\centering
			\centerline{\includegraphics[width=1.0\linewidth]{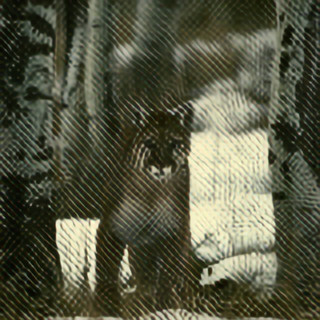}}
			\centerline{(e)}
		\end{minipage}
		\begin{minipage}[b]{0.16\linewidth}
			\centering
			\centerline{\includegraphics[width=1.0\linewidth]{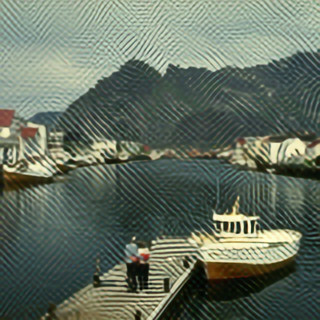}}
			\centerline{(f)}
		\end{minipage}
	\end{center}
	\caption{Visual examples of the universal attack with $\alpha=4/255$ on the BSD100 dataset \cite{martin2001database} for the RCAN model. (a) LR (original) (b) Perturbation (c) LR (attacked) (d) SR (e--f) Other examples obtained from the images attacked with the same perturbation}
	\label{fig:universal_example_rcan}
\end{figure*}

\noindent \textbf{Relation to model objectives.}
ESRGAN and 4PP-EUSR, which employ GANs for considering perceptual quality improvement, produce more significantly degraded outputs than the other methods.
Since ESRGAN has exactly the same structure as RRDB but is trained with a different objective (i.e., considering perceptual quality), the more significant vulnerability of ESRGAN than RRDB implies that differences of the training objectives affect the robustness against the adversarial attacks.
It is known that the methods employing GANs tend to generate sharper textures than the other methods to ensure naturally appealing quality of the upscaled images \cite{blau2018perception}.
Therefore, these methods amplify small perturbations significantly and produce undesirable textures, which makes them more vulnerable to the adversarial attacks than the methods without GANs.
\\[-0.8\baselineskip]

\noindent \textbf{Relation to model sizes.}
It is observed that the vulnerability of the super-resolution models is related to their model sizes.
For example, EDSR-baseline, which is a smaller version of EDSR, shows higher PSNR values for SR images than EDSR, as shown in Figures~\ref{fig:basic_alpha_psnr}d, \ref{fig:basic_alpha_psnr}e, and \ref{fig:basic_alpha_psnr}f.
This is confirmed in \figurename~\ref{fig:relation_model_size}, where we compare the robustness with respect to the model size.
The figure explains that the PSNR values of SR images tend to decrease when more model parameters or more convolutional layers are employed.
Further analysis on this phenomenon is given in Section~\ref{sec:partial_attack}.
\\[-0.8\baselineskip]

\begin{figure*}[t]
	\begin{center}
		\centering
		\begin{minipage}[b]{0.33\linewidth}
			\centering
			\centerline{\includegraphics[width=1.0\linewidth]{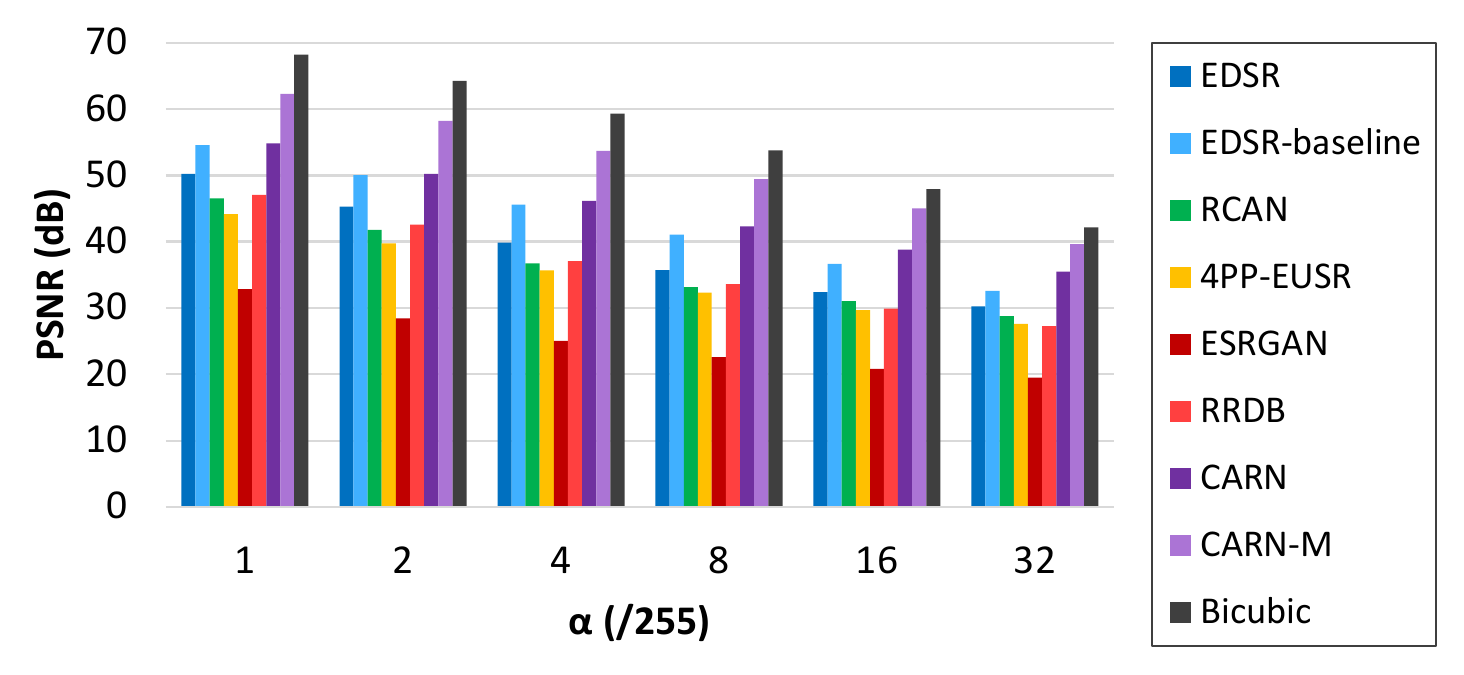}}
			\centerline{(a) Set5}
		\end{minipage}
		\begin{minipage}[b]{0.33\linewidth}
			\centering
			\centerline{\includegraphics[width=1.0\linewidth]{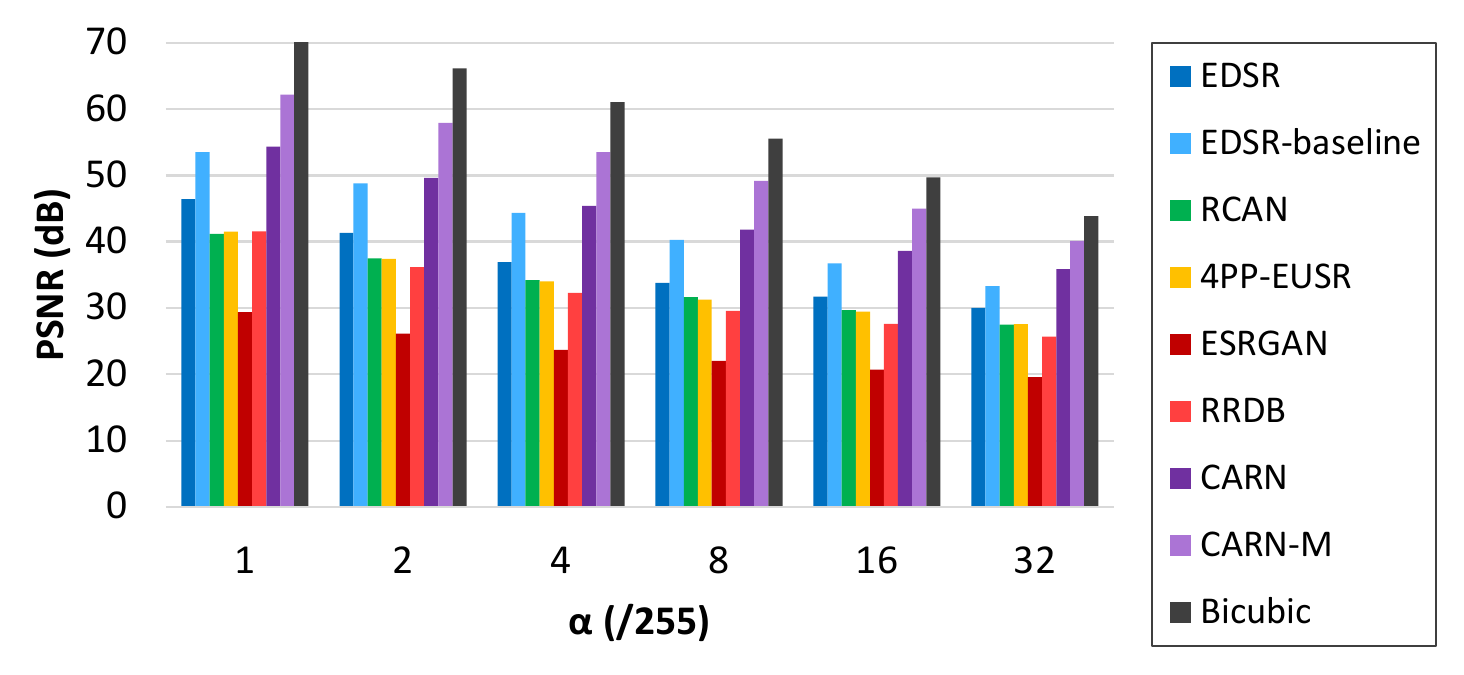}}
			\centerline{(b) Set14}
		\end{minipage}
		\begin{minipage}[b]{0.33\linewidth}
			\centering
			\centerline{\includegraphics[width=1.0\linewidth]{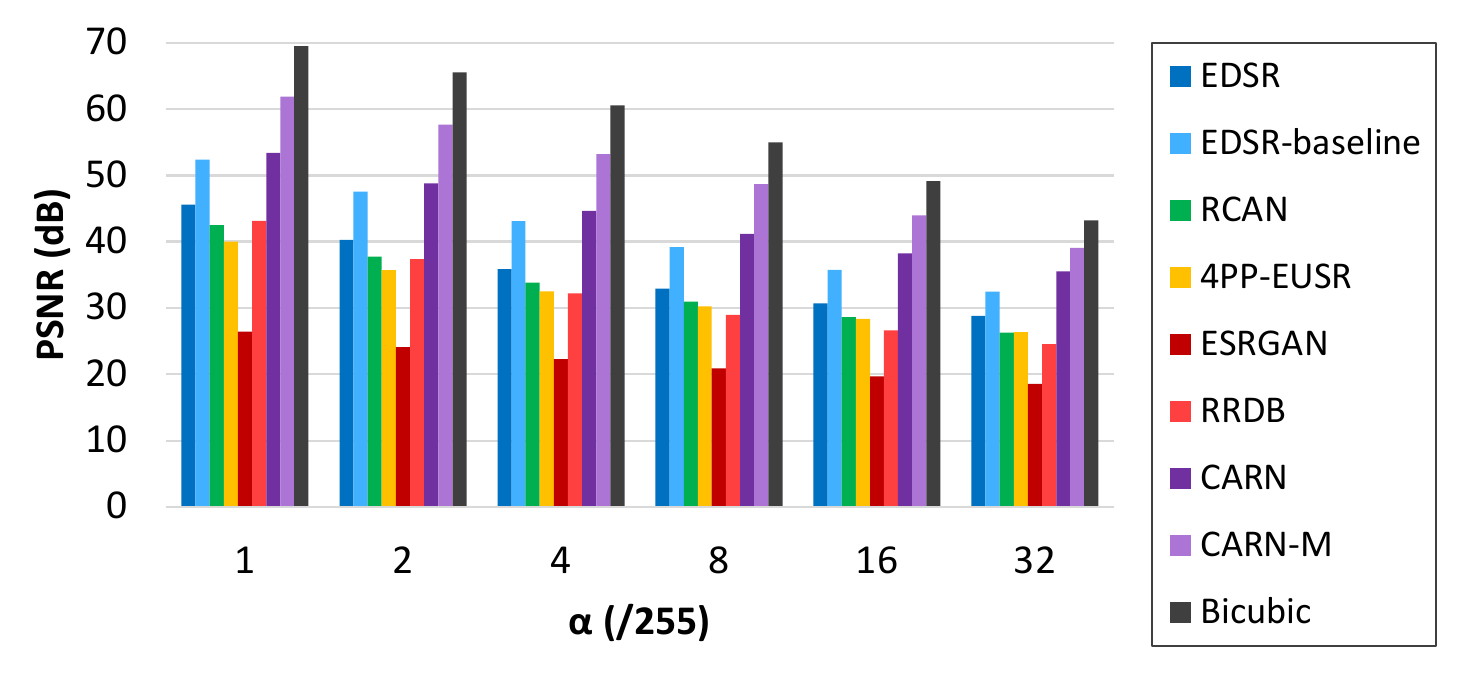}}
			\centerline{(c) BSD100}
		\end{minipage}
	\end{center}
	\caption{Comparison of the PSNR values of SR images with respect to different $\alpha$ values for the partial attack.}
	\label{fig:partial_alpha_psnr}
\end{figure*}

\begin{figure*}[t]
	\begin{center}
		\centering
		\begin{minipage}[b]{0.188\linewidth}
			\centering
			\centerline{\includegraphics[width=0.98\linewidth]{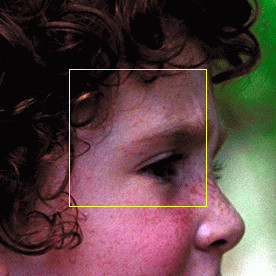}}
			\centerline{Ground-truth}
		\end{minipage}
		\begin{minipage}[b]{0.188\linewidth}
			\centering
			\centerline{\includegraphics[width=0.98\linewidth]{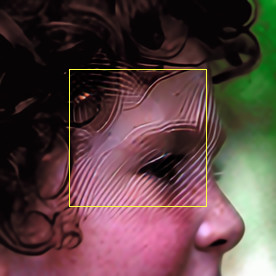}}
			\centerline{EDSR}
		\end{minipage}
		\begin{minipage}[b]{0.188\linewidth}
			\centering
			\centerline{\includegraphics[width=0.98\linewidth]{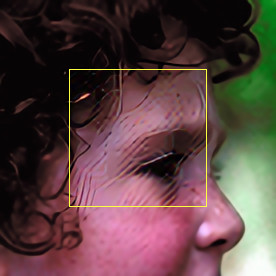}}
			\centerline{EDSR-baseline}
		\end{minipage}
		\begin{minipage}[b]{0.188\linewidth}
			\centering
			\centerline{\includegraphics[width=0.98\linewidth]{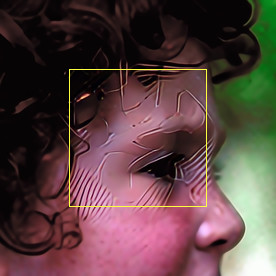}}
			\centerline{RCAN}
		\end{minipage}
		\begin{minipage}[b]{0.188\linewidth}
			\centering
			\centerline{\includegraphics[width=0.98\linewidth]{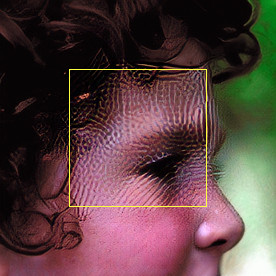}}
			\centerline{4PP-EUSR}
		\end{minipage}
		\\ \medskip
		\begin{minipage}[b]{0.188\linewidth}
			\centering
			\centerline{\includegraphics[width=0.98\linewidth]{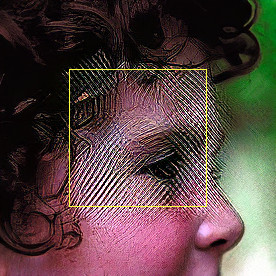}}
			\centerline{ESRGAN}
		\end{minipage}
		\begin{minipage}[b]{0.188\linewidth}
			\centering
			\centerline{\includegraphics[width=0.98\linewidth]{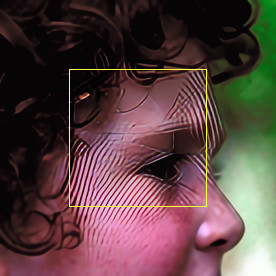}}
			\centerline{RRDB}
		\end{minipage}
		\begin{minipage}[b]{0.188\linewidth}
			\centering
			\centerline{\includegraphics[width=0.98\linewidth]{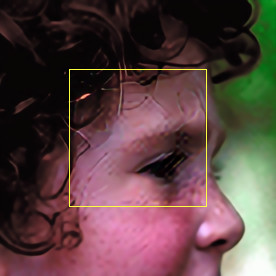}}
			\centerline{CARN}
		\end{minipage}
		\begin{minipage}[b]{0.188\linewidth}
			\centering
			\centerline{\includegraphics[width=0.98\linewidth]{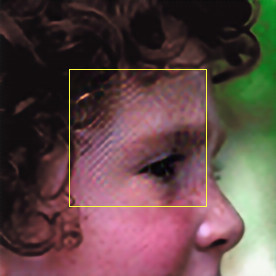}}
			\centerline{CARN-M}
		\end{minipage}
		\begin{minipage}[b]{0.188\linewidth}
			\centering
			\centerline{\includegraphics[width=0.98\linewidth]{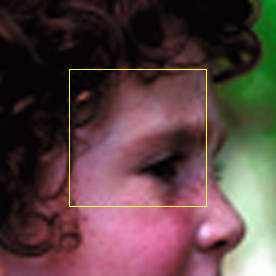}}
			\centerline{Bicubic}
		\end{minipage}
	\end{center}
	\caption{Visual comparison of the SR images for the partial adversarial attack with $\alpha=8/255$ on an image of Set14 \cite{zeyde2010single}. The regions marked with yellow boxes correspond to the regions where the attack is applied in the LR images.}
	\label{fig:partial_example}
\end{figure*}

\noindent \textbf{Transferability.}
In the classification tasks, the ``transferability'' means the possibility that a misclassified adversarial example is also misclassified by another classifier \cite{liu2016delving}.
We also examine the transferability of adversarial attacks in super-resolution.
In other words, an adversarial example that is found for a ``source'' super-resolution model is inputted to another ``target'' model, and the PSNR value of the output image is measured.

\figurename~\ref{fig:basic_transferability_psnr} summarizes the transferability for the deep learning-based super-resolution models on the BSD100 dataset, where $\alpha=8/255$.
%\footnote{See the Supplementary Material for visualized results.}.
The figure shows that the adversarial examples are transferable between different models to some extent, and the level of transferability differs depending on the combination of the source and target models.
The adversarial examples found for CARN and EDSR-baseline are highly transferable, while those for RCAN are the least transferable.
The result implies that RCAN has its own specific characteristics in recovering the textures from the input images, which makes the perturbations associated with such characteristics less effective in the other super-resolution methods.

\subsection{Universal attack}

\figurename~\ref{fig:universal_alpha_psnr} compares the performance of the super-resolution methods for the BSD100 dataset with respect to different $\alpha$ values when the universal attack is applied.
The figure confirms that the super-resolution models are also vulnerable to the image-agnostic universal attack, although the universal attack requires larger perturbations of the input images (i.e., slightly lower PSNR values in \figurename~\ref{fig:universal_alpha_psnr}a than in \figurename~\ref{fig:basic_alpha_psnr}c) and is slightly less powerful than the image-specific attack (i.e., slightly higher PSNR values in \figurename~\ref{fig:universal_alpha_psnr}b than in \figurename~\ref{fig:basic_alpha_psnr}f).
Compared to the results of the basic attack (\figurename~\ref{fig:basic_alpha_psnr}), the same tendency is observed: both ESRGAN and 4PP-EUSR are the most vulnerable and the bicubic interpolation is the most robust.

\figurename~\ref{fig:universal_example_rcan} shows visual examples of the universal attack for RCAN, where $\alpha$ is $4/255$.
From all images of the BSD100 dataset, our attack method finds a universal perturbation (\figurename~\ref{fig:universal_example_rcan}b), which changes the input image shown in \figurename~\ref{fig:universal_example_rcan}a to the one in \figurename~\ref{fig:universal_example_rcan}c.
While the attacked LR image has hardly noticeable differences from the original image, its upscaled version contains significant artifacts as shown in \figurename~\ref{fig:universal_example_rcan}d.
Similar artifacts can be observed in the other SR images attacked with the same perturbation, as shown in Figures~\ref{fig:universal_example_rcan}e and \ref{fig:universal_example_rcan}f.
This demonstrates that the state-of-the-art super-resolution methods using deep learning are also vulnerable to the universal perturbation.

\subsection{Partial attack}
\label{sec:partial_attack}

\figurename~\ref{fig:partial_alpha_psnr} shows the PSNR values of the SR images for the partial attack with respect to different $\alpha$ values.
The rank of the super-resolution methods in terms of PSNR is the same to that for the basic attack, except that the PSNR values of the partial attack are much higher than those of the basic attack, since the region where PSNR is measured is not directly perturbed in the LR image.
This shows that the propagation of the perturbation to the neighboring pixels during upscaling accounts for different levels of vulnerability of different super-resolution models.
For instance, all the PSNR values of ESRGAN, except for $\alpha=1$ in Set5, are lower than 30 dB due to the partial attack.

\figurename~\ref{fig:partial_example} shows example SR images obtained from an image in the Set14 dataset that are partially attacked with $\alpha = 8/255$.
The degradation due to the attack propagates outside of the attacked region, which are particularly noticeable for ESRGAN and RRDB.
This is because the kernels of the convolutional layers operate on not only the pixel of a target position but also its adjacent pixels.
Moreover, the propagation of the perturbation due to such operations is further extended through multiple convolutional layers, which accounts for the result shown in \figurename~\ref{fig:relation_model_size}b.

%-------------------------------------------------------------------------
\section{Advanced Topics}
\label{sec:advanced_topics}

\begin{figure}[t]
	\begin{center}
		\centering
		\begin{minipage}[b]{0.32\linewidth}
			\centering
			\centerline{\includegraphics[width=1.0\linewidth]{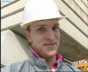}}
			\centerline{LR (original)}
		\end{minipage}
		\begin{minipage}[b]{0.32\linewidth}
			\centering
			\centerline{\includegraphics[width=1.0\linewidth]{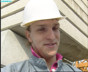}}
			\centerline{LR (target)}
		\end{minipage}
		\begin{minipage}[b]{0.32\linewidth}
			\centering
			\centerline{\includegraphics[width=1.0\linewidth]{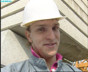}}
			\centerline{LR (attacked)}
		\end{minipage}
		\\ \medskip
		\begin{minipage}[b]{0.32\linewidth}
			\centering
			\centerline{\includegraphics[width=1.0\linewidth]{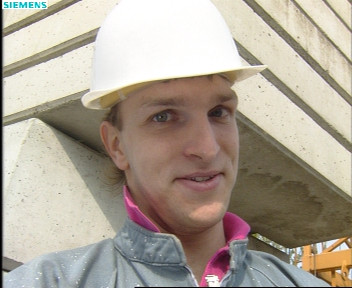}}
			\centerline{HR (original)}
		\end{minipage}
		\begin{minipage}[b]{0.32\linewidth}
			\centering
			\centerline{\includegraphics[width=1.0\linewidth]{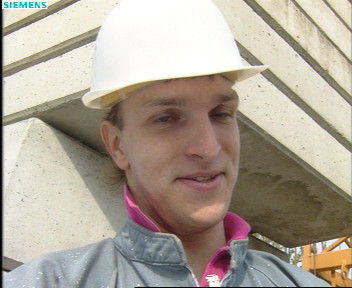}}
			\centerline{HR (target)}
		\end{minipage}
		\begin{minipage}[b]{0.32\linewidth}
			\centering
			\centerline{\includegraphics[width=1.0\linewidth]{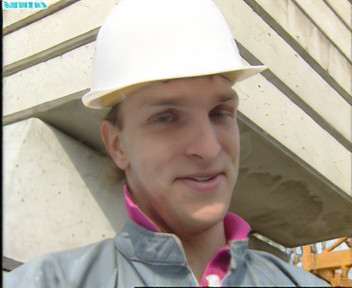}}
			\centerline{SR (attacked)}
		\end{minipage}
	\end{center}
	\caption{Result of the targeted attack with $\alpha=16/255$ using two frames of a video ``foreman'' \cite{video2008yuv} for 4PP-EUSR \cite{choi2018deep}.}
	\label{fig:targeted_attack_showcase}
\end{figure}

\subsection{Targeted attack}

In the case of the classification tasks, it is possible to attack an image so that a classifier wrongly classifies the image as a specific target class.
We present a showcase demonstrating that this concept can be also applied to the super-resolution methods.
In other words, instead of degrading quality of the output image, the targeted attack makes a super-resolution method generate an image that is more similar to a target image than the original ground-truth one.
For this, we modify (\ref{eq:basic_attack_iteration_tilde}) as:
\begin{equation}
\widetilde{\mathbf{X}}_{n+1} = \mathrm{clip}_{0, 1} \Big( \mathbf{X}_{n} - \frac{\alpha}{T}~\mathrm{sgn} \big( \nabla \mathcal{L} ( \mathbf{X}_{n}, \mathbf{X}^{*} ) \big) \Big)
\end{equation}
where $\mathbf{X}^{*}$ is the target image.

For demonstration, we use two adjacent frames of a video named ``foreman'' \cite{video2008yuv}.
\figurename~\ref{fig:targeted_attack_showcase} shows the result for 4PP-EUSR, where $\alpha = 16/255$ and $T=50$.
The figure shows that the targeted attack is successful: the perturbation is generated so as to make the super-resolution method produce the upscaled output (``SR (attacked)''), which looks more similar to the target high-resolution (HR) image with \textit{half-closed eyes} (``HR (target)'') than the original ground-truth image with \textit{open eyes} (``HR (original)''), while the attacked input image (``LR (attacked)'') still looks more similar to the original image (``LR (original)'') than the low-resolution version of the target image (``LR (target)'').
In addition, we conduct a subjective test with 20 human observers, and 10 of them recognized the attacked output (``SR (attacked)'') as \textit{closed eyes}.
These results have serious security implications: attacks on super-resolution can not only compromise the fundamental goal of super-resolution (i.e., image quality enhancement) but also jeopardize further manual or automatic examination of the super-resolved images (e.g., identifying persons or objects in surveillance cameras, recognizing text in images, etc.).

\subsection{Robustness measure}
\label{sec:robustness_measure}

\begin{figure}[t]
	\begin{center}
		\centering
		\includegraphics[width=0.99\linewidth]{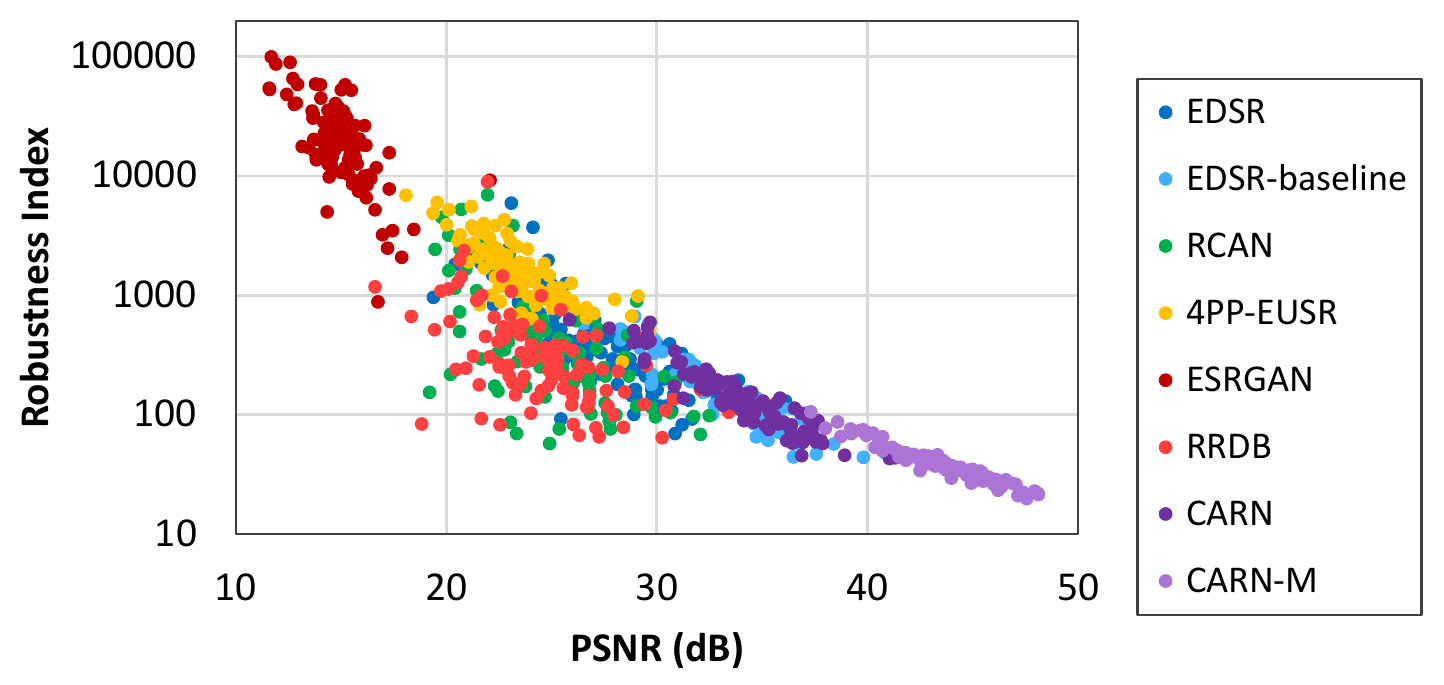}
	\end{center}
	\caption{PSNR vs. the robustness index for the BSD100 dataset \cite{martin2001database} when $\alpha=1/255$. Each point corresponds to each image in the dataset.}
	\label{fig:psnr_robustness}
\end{figure}

Recently, Weng \textit{et al.} \cite{weng2018evaluating} propose an \textit{attack-agnostic} robustness measure of classification models, called cross Lipschitz extreme value for network robustness (CLEVER), which does not depend on specific attack methods.
It estimates the lower bound of robustness using the cross Lipschitz constant based on the extreme value theory.
We apply the core idea of this method to the super-resolution tasks in order to theoretically validate the experimental results shown in Section~\ref{sec:experiments}.

Let $\mathbf{X}_{0}$ denote the original input image.
We first obtain ${N}_{s}$ random perturbations, which are within $[-\alpha, \alpha]$ for each pixel.
Let ${\Delta}^{(i)}$ denote the $i$-th random perturbation.
Then, we compute ${b}_{i} = || \nabla \mathcal{L} ( \mathbf{X}_{0} + {\Delta}^{(i)}, \mathbf{X}_{0} )  ||_{1}$ for all perturbations, where $\mathcal{L}$ is defined in (\ref{eq:basic_attack_sr_loss}).
Finally, we regard the maximum ${b}_{i}$ as the robustness index; a large robustness index indicates high vulnerability.
We set ${N}_{s}$ and $\alpha$ to 1024 and $1/255$, respectively.
%\footnote{See the Supplementary Material for the results using other $\alpha$ values.}.

\figurename~\ref{fig:psnr_robustness} shows the PSNR values for SR images and robustness indices of the eight deep learning-based super-resolution methods for the BSD100 dataset, where the PSNR values are obtained from the basic attack with the same $\alpha$ value (Section~\ref{sec:basic_attack_result}).
The result shows that the robustness index is strongly correlated to PSNR.
For instance, ESRGAN has the largest robustness indices, which shows the lowest PSNR values; the EDSR-baseline model has the similar robustness as the CARN model in terms of both PSNR  and the robustness index.
Furthermore, in each method, the robustness index successfully explains relative vulnerability of different images.
The applicability of the CLEVER method for explaining the robustness of the super-resolution methods implies that the underlying mechanisms of the adversarial attacks share similarity between the classification and super-resolution tasks.

\subsection{Defense}

We show two simple defense methods against attacks.
First, we adopt a resizing method \cite{xie2017mitigating} by reducing the size of the attacked input image by one pixel and then resizing it back to the original resolution, which is then inputted to the SR model.
With this, PSNR for EDSR with $\alpha = 8/255$ increases from 16.14
to 25.01 dB.
Second, we employ the geometric self-ensemble method used in the EDSR model \cite{lim2017enhanced}.
With this, PSNR for EDSR with $\alpha = 8/255$ increases from 16.14
to 23.47 dB.
More advanced defense methods can be investigated in the future work.

%------------------------------------------------------------------------
\section{Conclusion}
\label{sec:conclusion}

We have investigated the robustness of deep learning-based super-resolution methods against adversarial attacks, for which the attack methods for the classification tasks are optimized for our objectives.
Our results showed that state-of-the-art deep learning-based super-resolution methods are highly vulnerable to adversarial attacks, which is largely due to the perturbation propagation through the convolutional operation.
It was possible to measure different levels of robustness of different methods using the attack-agnostic robustness measure.
We also showed the feasibility of generating universal attacks and transferring attacks across super-resolution methods.
Furthermore, it was shown that the targeted attack can change the content of the image during super-resolution.

%------------------------------------------------------------------------
\section*{Acknowledgement}
This research was supported by the MSIT (Ministry of Science and ICT), Korea, under the ``ICT Consilience Creative Program'' (IITP-2019-2017-0-01015) supervised by the IITP (Institute for Information \& Communications Technology Planning \& Evaluation). In addition, this work was also supported by the IITP grant funded by the Korea government (MSIT) (R7124-16-0004, Development of Intelligent Interaction Technology Based on Context Awareness and Human Intention Understanding).

{\small
	\bibliographystyle{ieee_fullname}
	\bibliography{egbib}
}

\newpage

%------------------------------------------------------------------------
\section*{Supplementary Material}
In this supplementary material, we provide additional results that could not be included in the main paper due to the page limit.
\\[-0.7\baselineskip]

\begin{figure*}[t]
	\vspace{12pt}
	\begin{center}
		\centering
		\begin{minipage}[b]{0.195\linewidth}
			\centering
			\centerline{\includegraphics[width=0.98\linewidth]{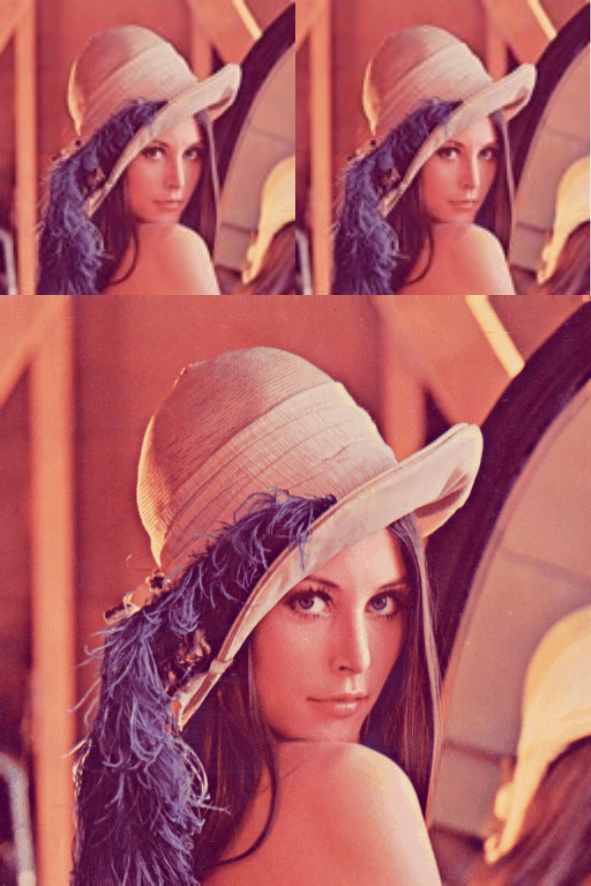}}
			\centerline{Ground-truth}
		\end{minipage}
		\begin{minipage}[b]{0.195\linewidth}
			\centering
			\centerline{\includegraphics[width=0.98\linewidth]{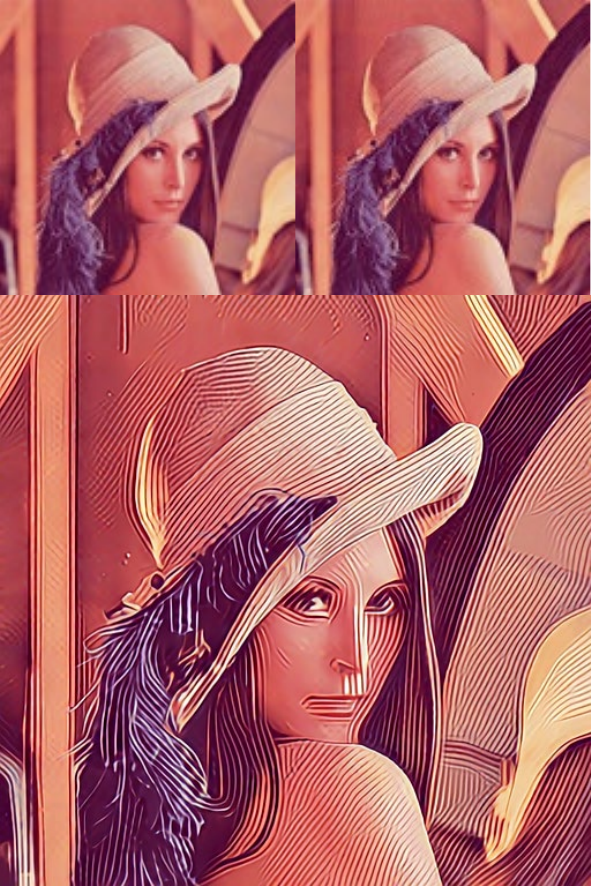}}
			\centerline{EDSR}
		\end{minipage}
		\begin{minipage}[b]{0.195\linewidth}
			\centering
			\centerline{\includegraphics[width=0.98\linewidth]{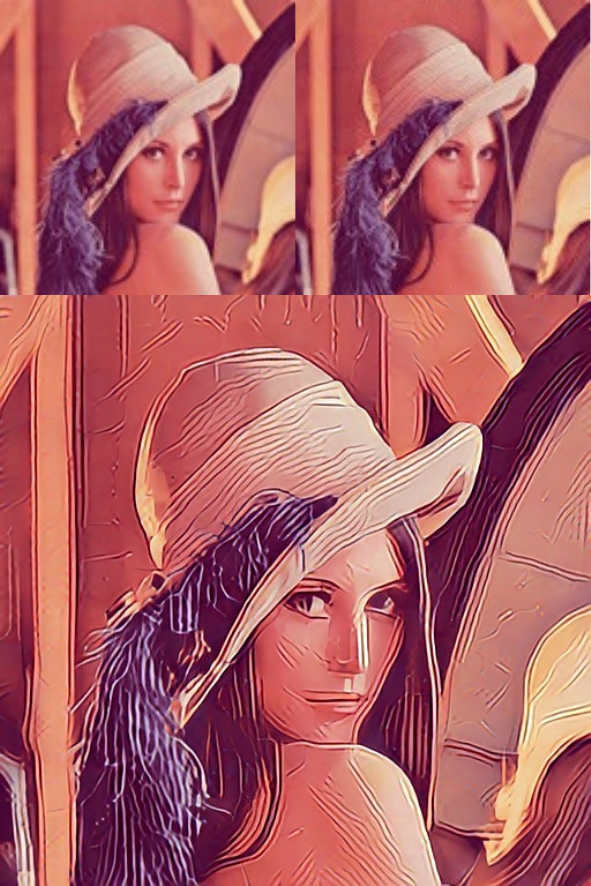}}
			\centerline{EDSR-baseline}
		\end{minipage}
		\begin{minipage}[b]{0.195\linewidth}
			\centering
			\centerline{\includegraphics[width=0.98\linewidth]{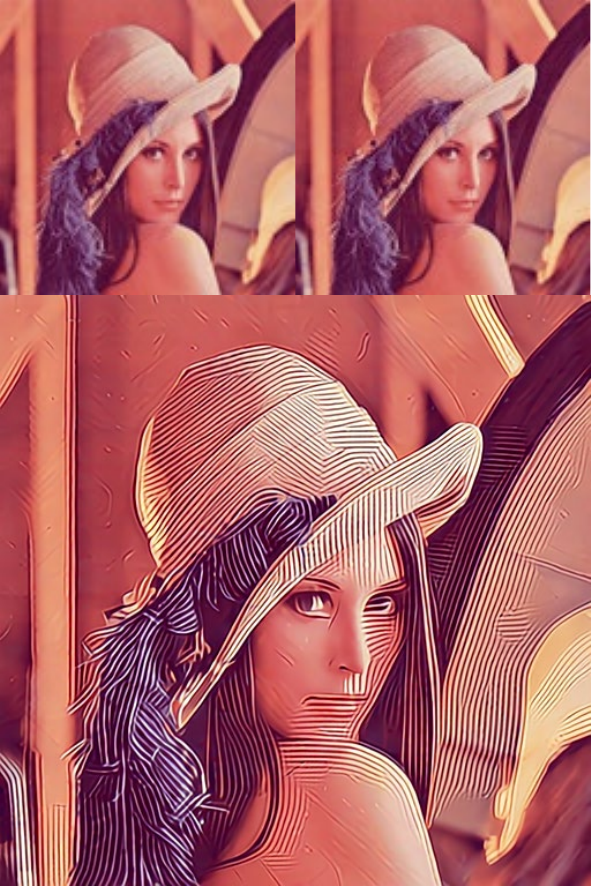}}
			\centerline{RCAN}
		\end{minipage}
		\begin{minipage}[b]{0.195\linewidth}
			\centering
			\centerline{\includegraphics[width=0.98\linewidth]{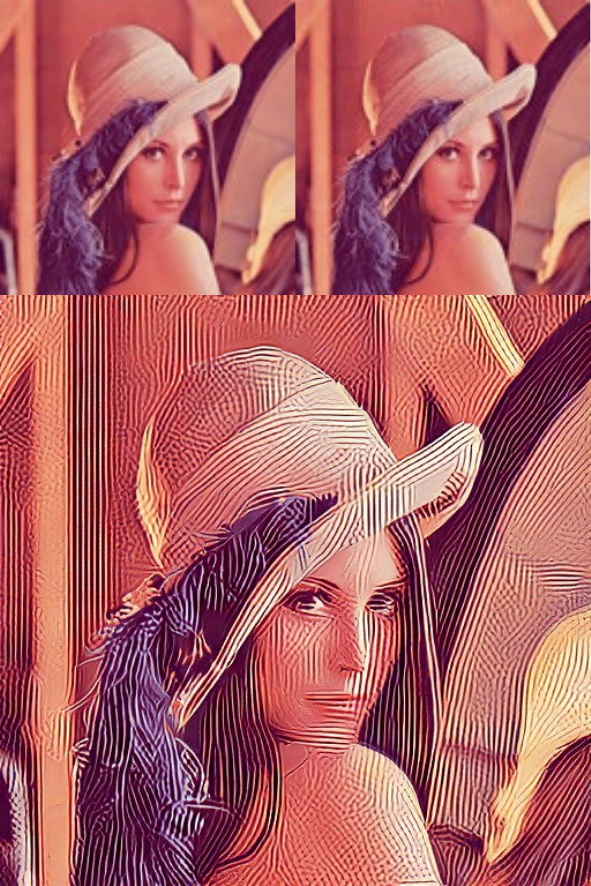}}
			\centerline{4PP-EUSR}
		\end{minipage}
		\\ \medskip
		\begin{minipage}[b]{0.195\linewidth}
			\centering
			\centerline{\includegraphics[width=0.98\linewidth]{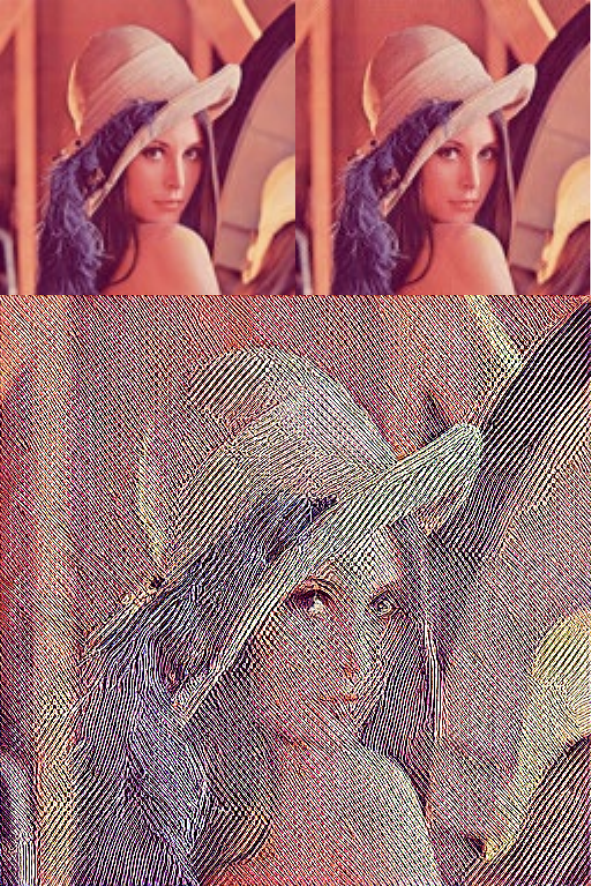}}
			\centerline{ESRGAN}
		\end{minipage}
		\begin{minipage}[b]{0.195\linewidth}
			\centering
			\centerline{\includegraphics[width=0.98\linewidth]{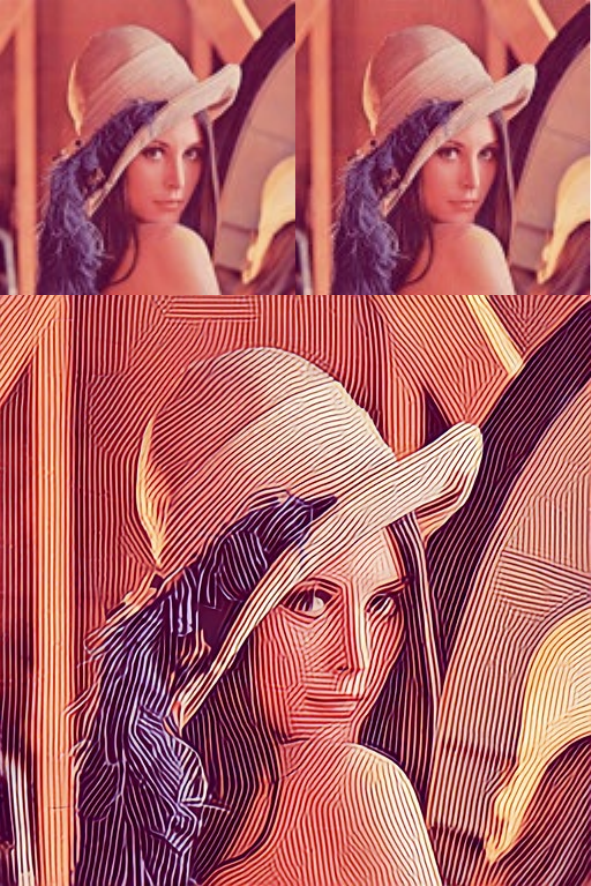}}
			\centerline{RRDB}
		\end{minipage}
		\begin{minipage}[b]{0.195\linewidth}
			\centering
			\centerline{\includegraphics[width=0.98\linewidth]{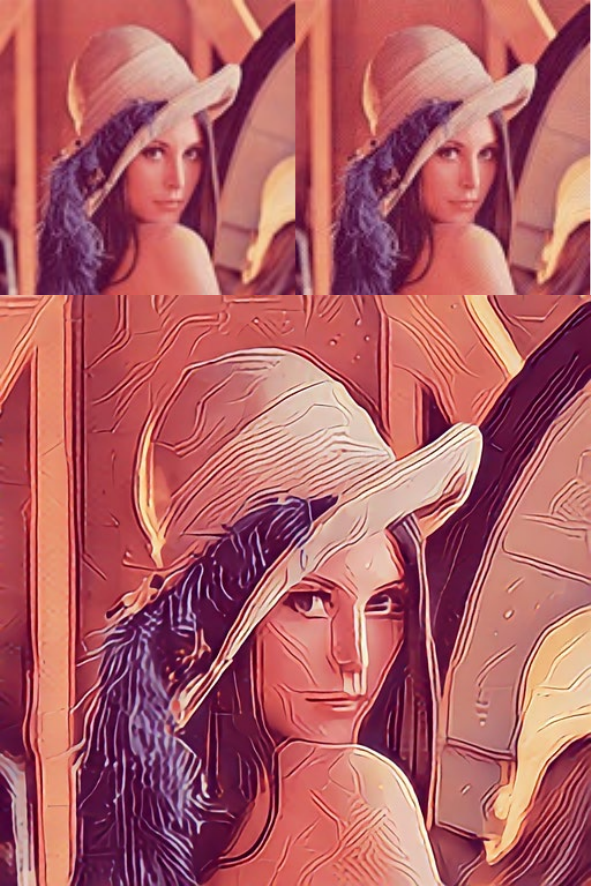}}
			\centerline{CARN}
		\end{minipage}
		\begin{minipage}[b]{0.195\linewidth}
			\centering
			\centerline{\includegraphics[width=0.98\linewidth]{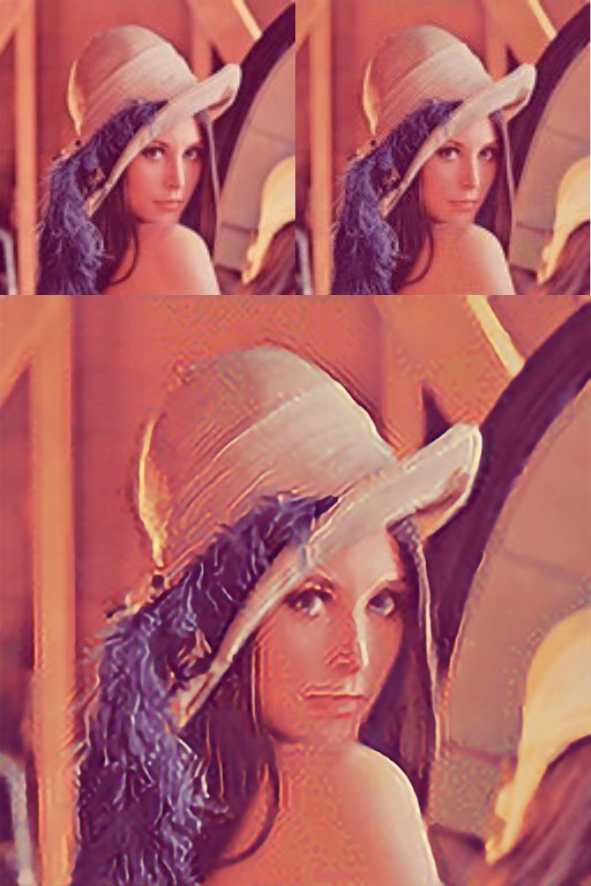}}
			\centerline{CARN-M}
		\end{minipage}
		\begin{minipage}[b]{0.195\linewidth}
			\centering
			\centerline{\includegraphics[width=0.98\linewidth]{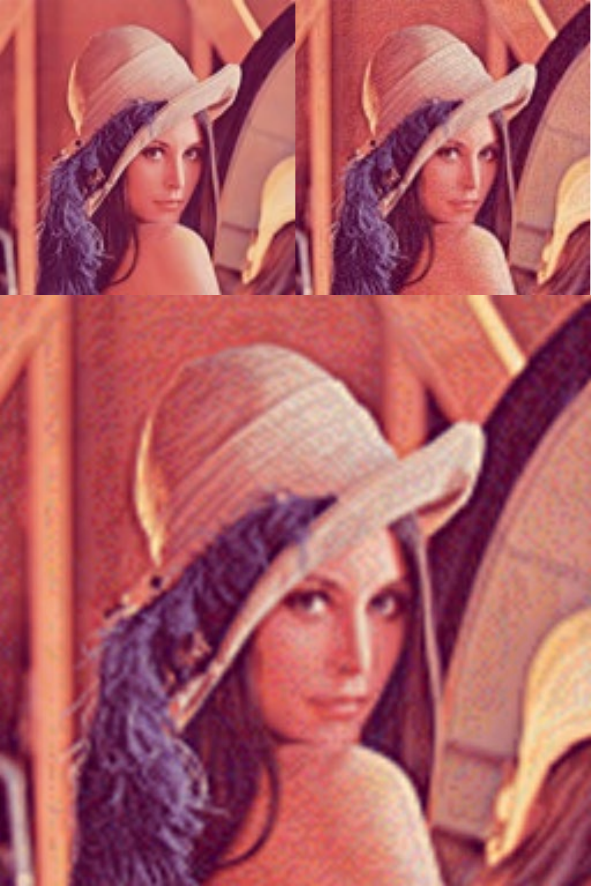}}
			\centerline{Bicubic}
		\end{minipage}
	\end{center}
	\caption{Visual comparison of the super-resolved outputs for the inputs attacked with $\alpha=8/255$. In each case, (top-left) is the original input in Set14 \cite{zeyde2010single}, (top-right) is the adversarial input, and (bottom) is the output obtained from the adversarial input. The input images are enlarged two times for better visualization.}
	\label{fig:basic_example_sm}
	\vspace{12pt}
\end{figure*}

\begin{figure*}[t]
	\vspace{12pt}
	\begin{center}
		\centering
		\begin{minipage}[b]{0.195\linewidth}
			\centering
			\centerline{\includegraphics[width=0.98\linewidth]{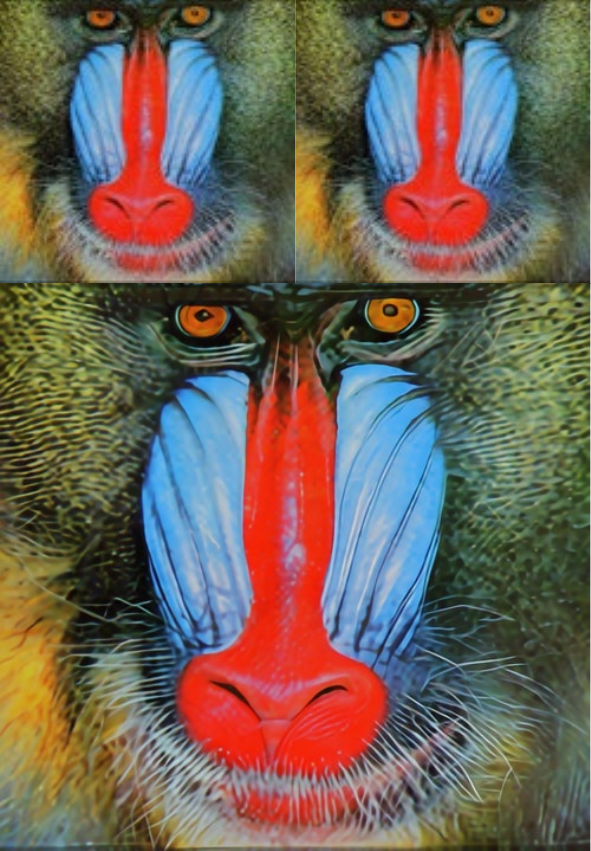}}
			\centerline{$\alpha=1/255$}
		\end{minipage}
		\begin{minipage}[b]{0.195\linewidth}
			\centering
			\centerline{\includegraphics[width=0.98\linewidth]{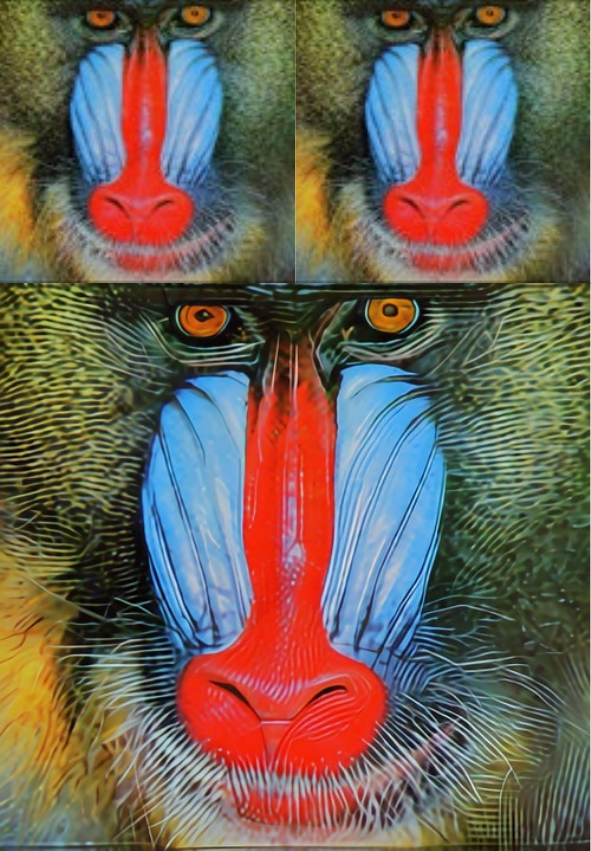}}
			\centerline{$\alpha=2/255$}
		\end{minipage}
		\begin{minipage}[b]{0.195\linewidth}
			\centering
			\centerline{\includegraphics[width=0.98\linewidth]{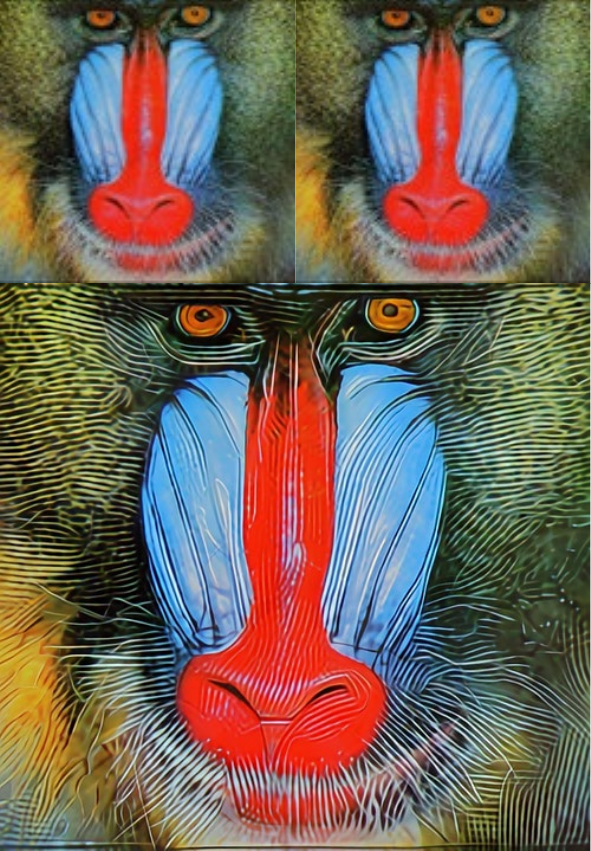}}
			\centerline{$\alpha=4/255$}
		\end{minipage}
		\begin{minipage}[b]{0.195\linewidth}
			\centering
			\centerline{\includegraphics[width=0.98\linewidth]{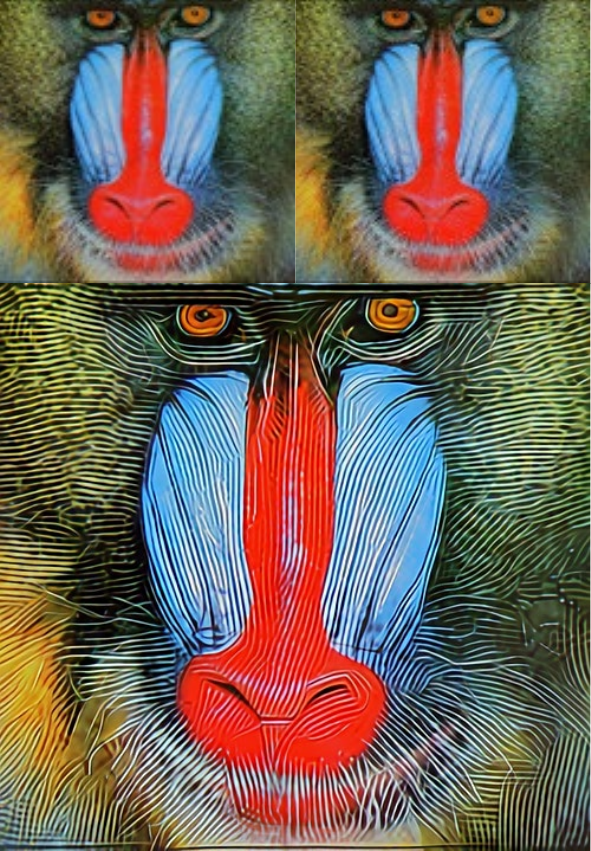}}
			\centerline{$\alpha=8/255$}
		\end{minipage}
		\begin{minipage}[b]{0.195\linewidth}
			\centering
			\centerline{\includegraphics[width=0.98\linewidth]{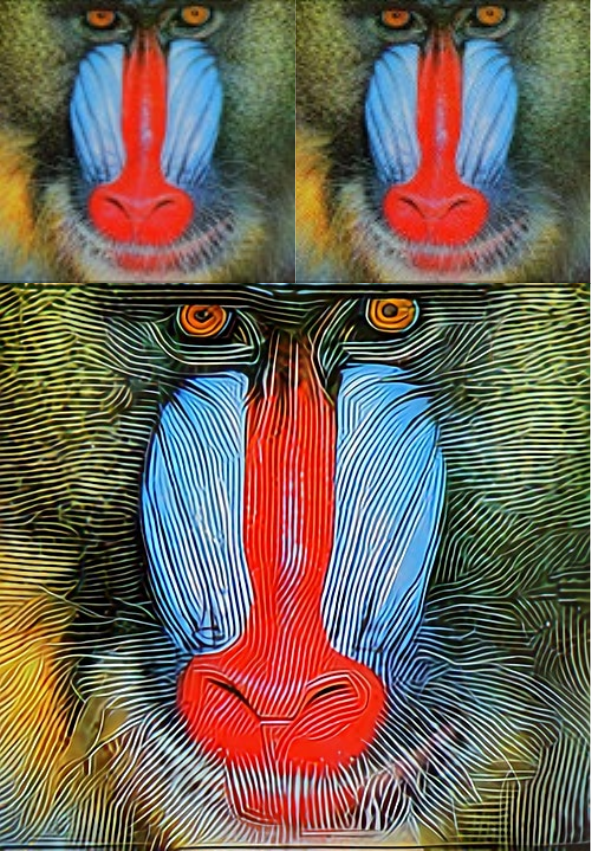}}
			\centerline{$\alpha=16/255$}
		\end{minipage}
	\end{center}
	\caption{Visual comparison of the super-resolved outputs for the inputs attacked with different $\alpha$ values. In each case, (top-left) is the original input in Set14 \cite{zeyde2010single}, (top-right) is the adversarial input, and (bottom) is the output obtained on EDSR \cite{lim2017enhanced}. The input images are enlarged two times for better visualization.}
	\label{fig:basic_example_diffalpha}
	\vspace{12pt}
\end{figure*}

\noindent \textbf{More visual comparisons of the basic attack.}
We provide two additional visual comparisons of the basic attack shown in Section 4.1 of the main paper.
\figurename~\ref{fig:basic_example_sm} shows additional example low-resolution (LR) and super-resolved (SR) images obtained from an image of Set14 \cite{zeyde2010single} with $\alpha=8/255$.
Undesirable artifacts similar to those observed in Figure 2 of the main paper can be found.
\figurename~\ref{fig:basic_example_diffalpha} shows the example images obtained by the EDSR model \cite{lim2017enhanced} with different $\alpha$ values.
As $\alpha$ increases, the upscaled images become more deteriorated, whereas the perturbed input images still look similar to the original image.
The results support that the deep super-resolution methods are highly vulnerable against the adversarial attack in various cases.
\\[-0.7\baselineskip]

\begin{figure*}[t]
	\begin{center}
		\centering
		\includegraphics[width=0.82\linewidth]{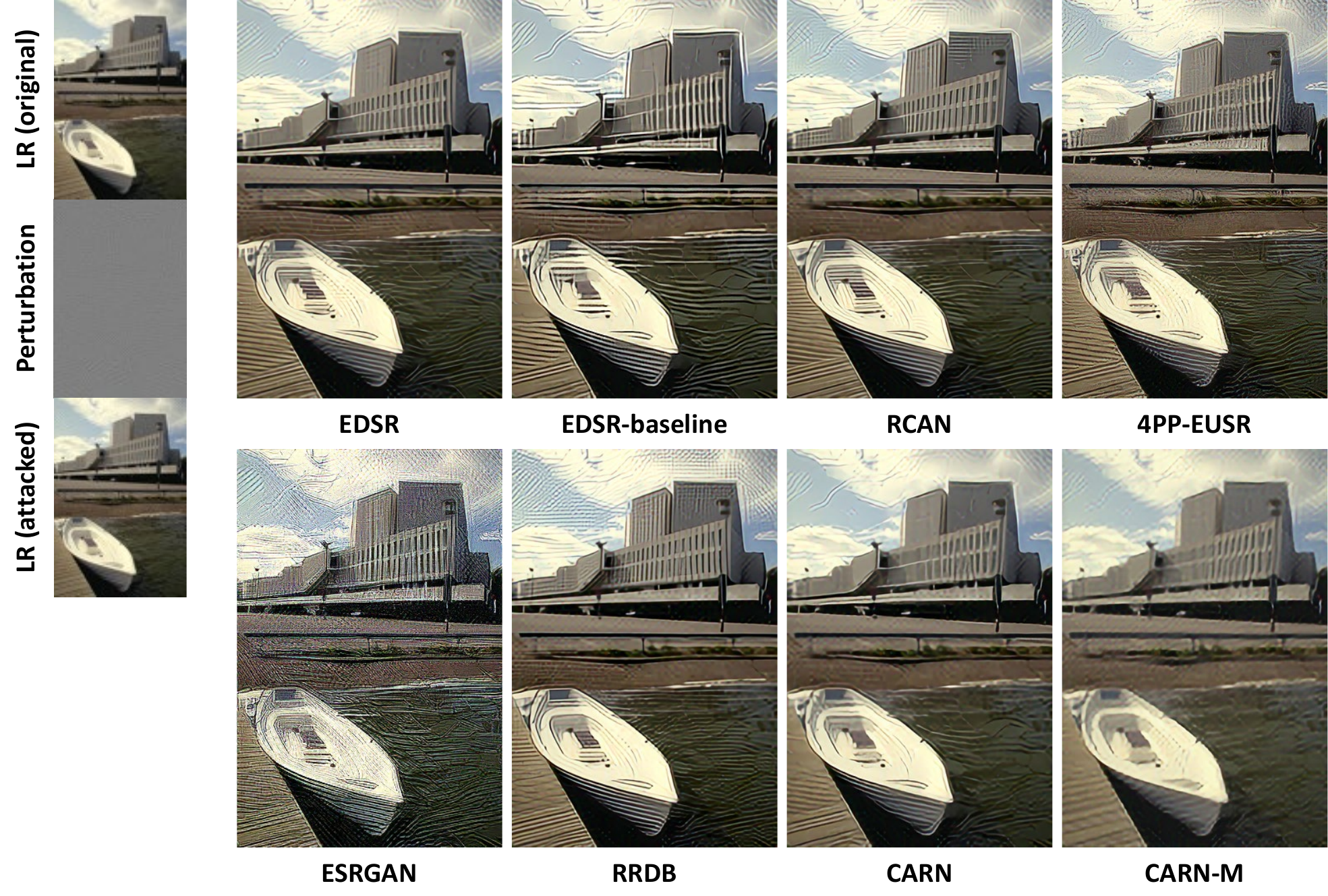}
	\end{center}
	\caption{Visual examples of the transferred attack where EDSR-baseline \cite{lim2017enhanced} is used as the source super-resolution model with $\alpha=8/255$. An image in the BSD100 \cite{martin2001database} dataset is used.}
	\label{fig:transferability_78004_edsr_baseline}
\end{figure*}

\begin{figure*}[t]
	\begin{center}
		\centering
		\includegraphics[width=0.82\linewidth]{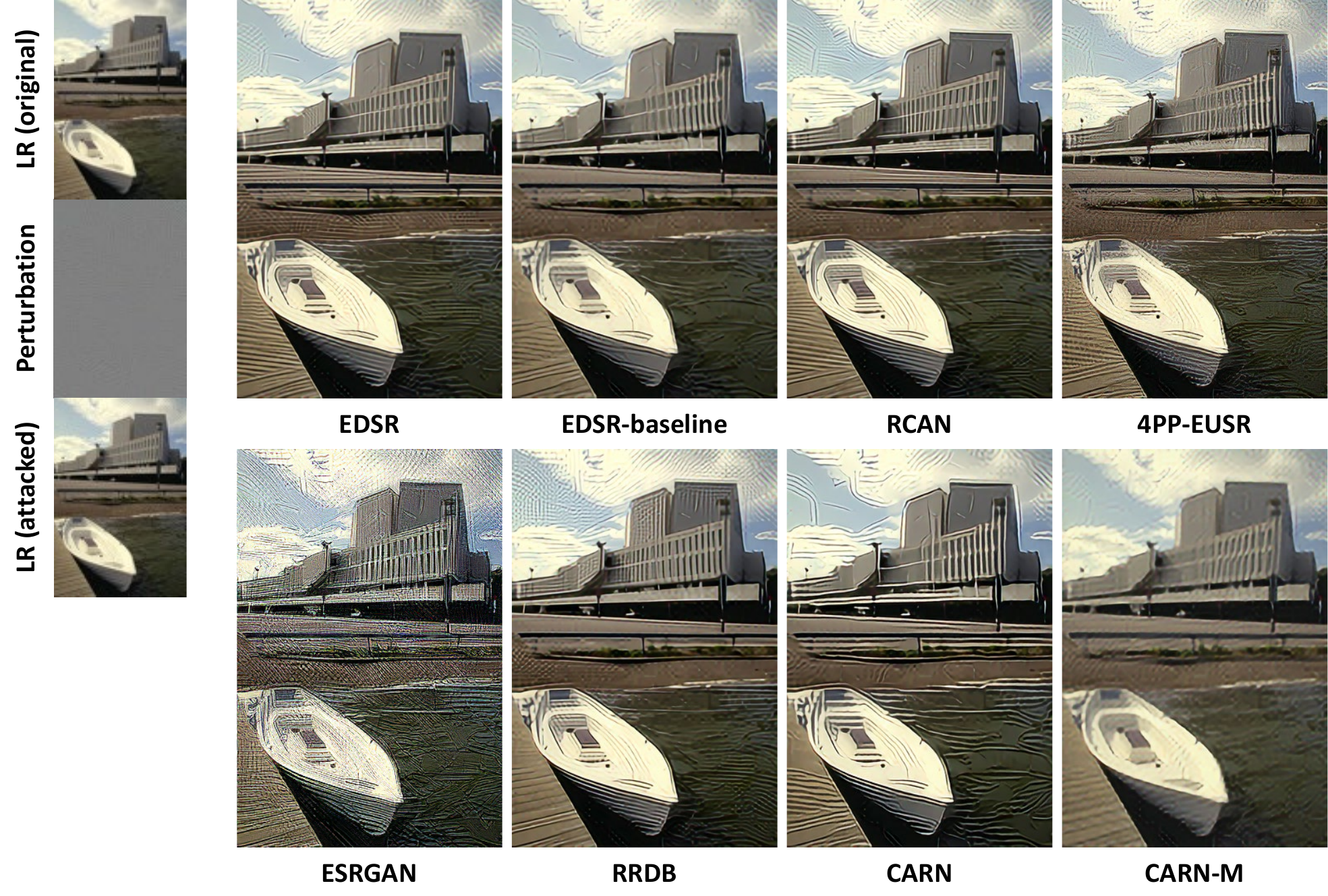}
	\end{center}
	\caption{Visual examples of the transferred attack where CARN \cite{ahn2018fast} is used as the source super-resolution model with $\alpha=8/255$. An image in the BSD100 \cite{martin2001database} dataset is used.}
	\label{fig:transferability_78004_carn}
\end{figure*}

\begin{figure*}[t]
	\begin{center}
		\centering
		\includegraphics[width=0.82\linewidth]{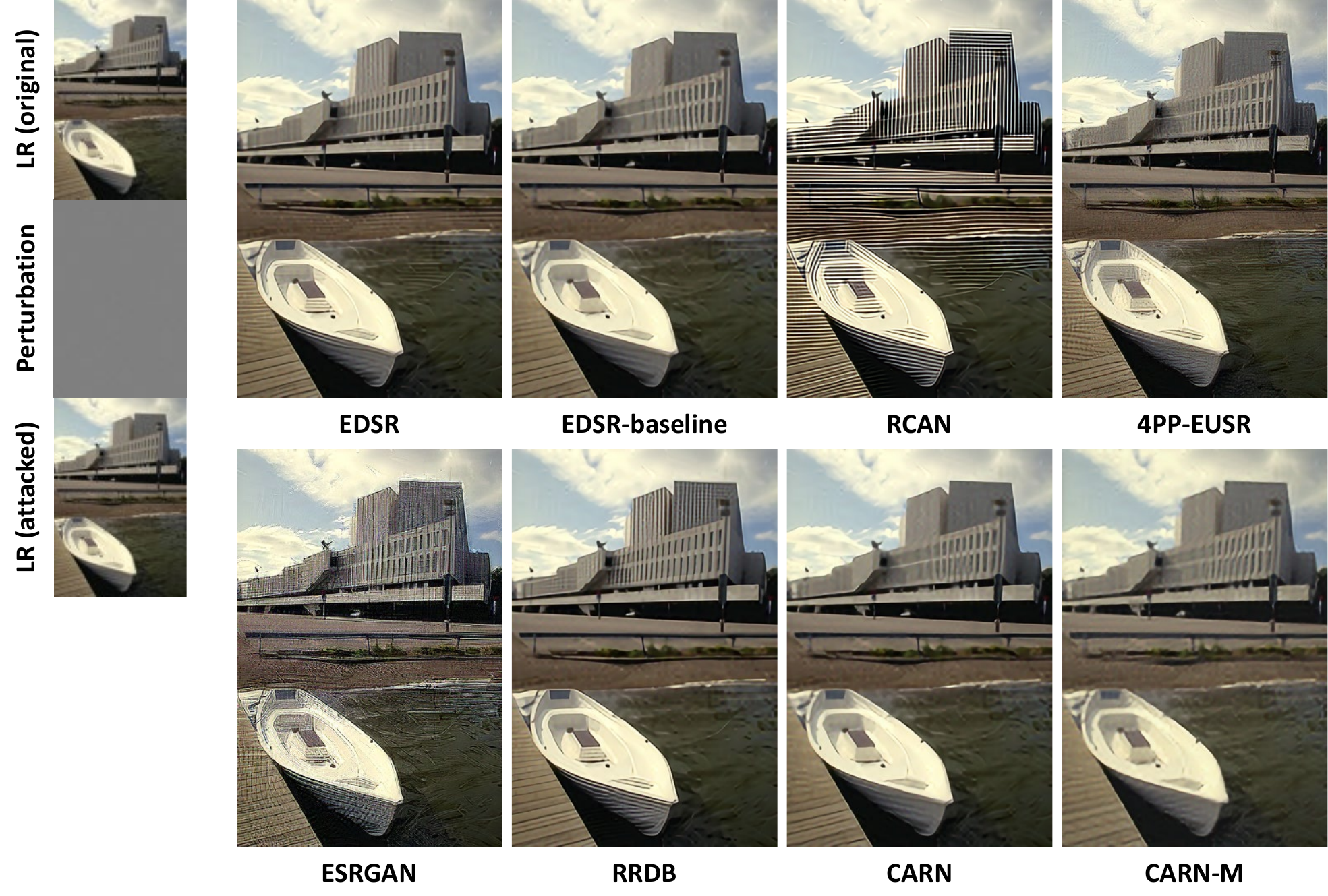}
	\end{center}
	\caption{Visual examples of the transferred attack where RCAN \cite{zhang2018image} is used as the source super-resolution model with $\alpha=8/255$. An image in the BSD100 \cite{martin2001database} dataset is used.}
	\label{fig:transferability_78004_rcan}
\end{figure*}

\begin{figure*}[t]
	\begin{center}
		\centering
		\includegraphics[width=0.82\linewidth]{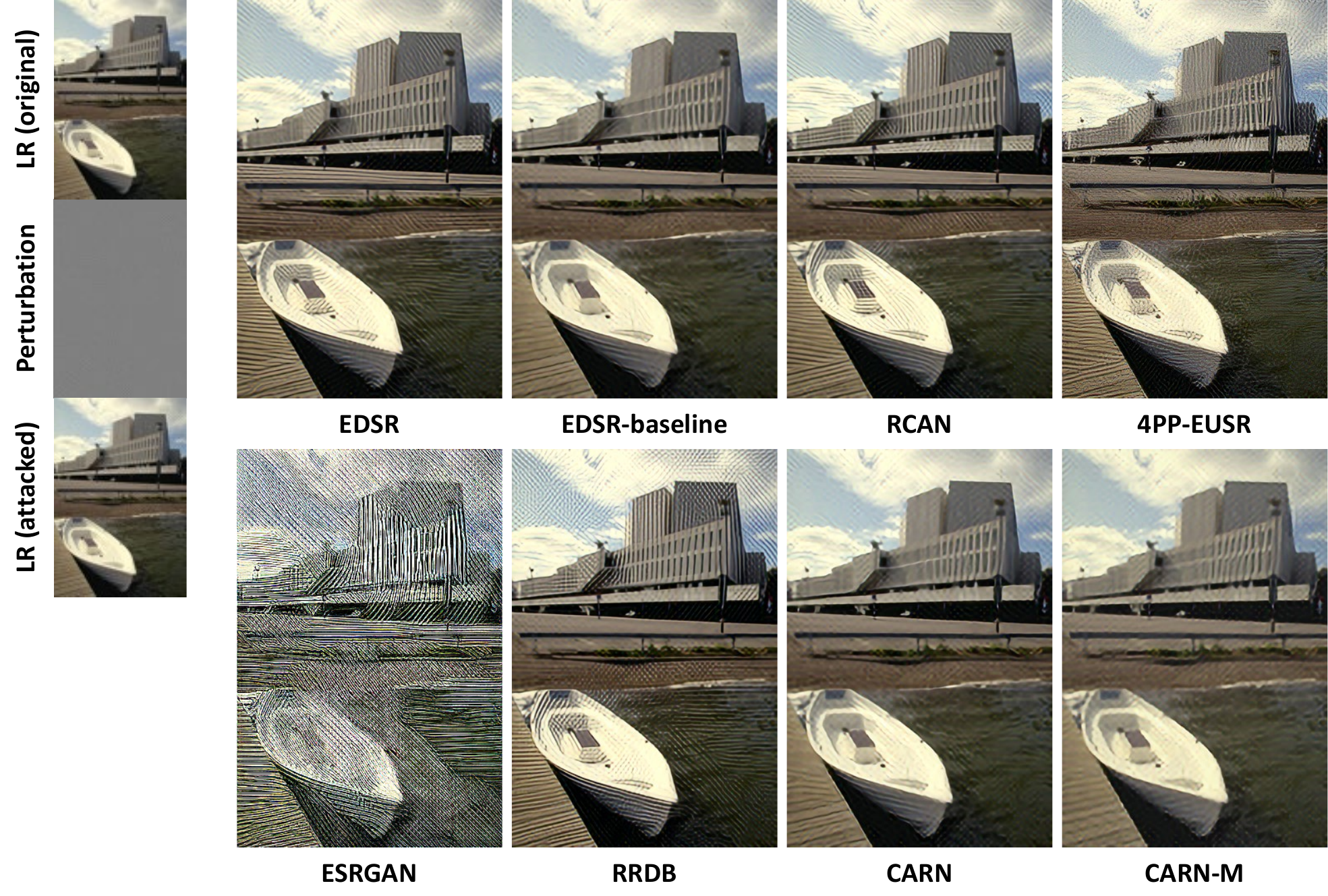}
	\end{center}
	\caption{Visual examples of the transferred attack where ESRGAN \cite{wang2018esrgan} is used as the source super-resolution model with $\alpha=8/255$. An image in the BSD100 \cite{martin2001database} dataset is used.}
	\label{fig:transferability_78004_esrgan}
\end{figure*}

\noindent \textbf{Visualized results of transferability.}
In Section 4.1 of the main paper, we compared the transferability of the deep super-resolution methods in terms of peak signal-to-noise ratio (PSNR).
According to Figure~4 in the main paper, EDSR-baseline \cite{lim2017enhanced} and CARN \cite{ahn2018fast} show higher transferability than the other models, whereas RCAN \cite{zhang2018image} and ESRGAN \cite{wang2018esrgan} show lower transferability.
Here, we visually explain the transferability of these four super-resolution methods in Figures \ref{fig:transferability_78004_edsr_baseline}, \ref{fig:transferability_78004_carn}, \ref{fig:transferability_78004_rcan}, and \ref{fig:transferability_78004_esrgan}.
In the figures, a LR image in the BSD100 dataset \cite{martin2001database} is attacked with one of the super-resolution models and inputted to the other super-resolution models including EDSR \cite{lim2017enhanced}, EDSR-baseline, RCAN, 4PP-EUSR \cite{choi2018deep}, ESRGAN, RRDB \cite{wang2018esrgan}, CARN, and CARN-M \cite{ahn2018fast}.
In Figures~\ref{fig:transferability_78004_edsr_baseline} and \ref{fig:transferability_78004_carn}, the attacked LR image successfully deteriorates the SR images obtained from the other methods, where similar fingerprint-like textures are observed as in Figure 2 of the main paper.
On the other hand, in Figures~\ref{fig:transferability_78004_rcan} and \ref{fig:transferability_78004_esrgan}, the perturbations found for RCAN and ESRGAN are not so effective for the other models; the amounts of deterioration in the SR images produced by the other models are much smaller than those triggered by the perturbations for EDSR-baseline and CARN (Figures~\ref{fig:transferability_78004_edsr_baseline} and \ref{fig:transferability_78004_carn}).
\\[-0.7\baselineskip]

\begin{figure*}[t]
	\begin{center}
		\centering
		\begin{minipage}[b]{0.20\linewidth}
			\centering
			\centerline{\includegraphics[width=0.98\linewidth]{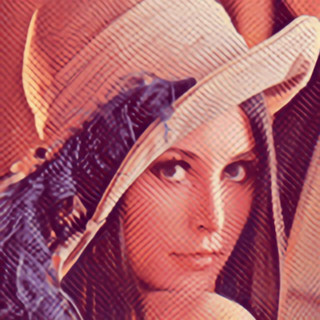}}
		\end{minipage}
		\begin{minipage}[b]{0.20\linewidth}
			\centering
			\centerline{\includegraphics[width=0.98\linewidth]{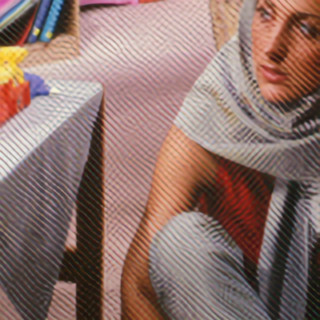}}
		\end{minipage}
		\begin{minipage}[b]{0.20\linewidth}
			\centering
			\centerline{\includegraphics[width=0.98\linewidth]{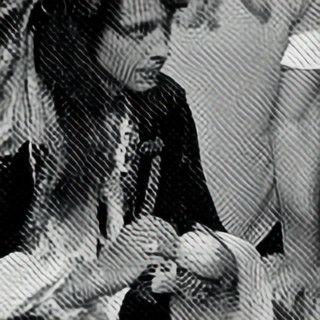}}
		\end{minipage}
		\begin{minipage}[b]{0.20\linewidth}
			\centering
			\centerline{\includegraphics[width=0.98\linewidth]{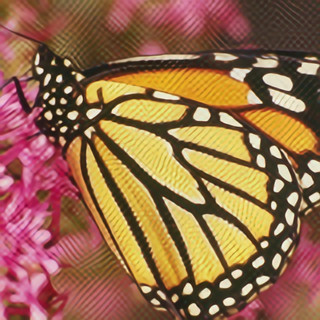}}
		\end{minipage}
	\end{center}
	\caption{Results of the universal attack applied to the Set14 dataset \cite{zeyde2010single}.}
	\label{fig:universal_transfer_set14}
	\vspace{16pt}
\end{figure*}

\noindent \textbf{Transferability of the universal attack.}
We examine the universal attack across datasets, i.e., the universal perturbation obtained for the BSD100 dataset \cite{martin2001database} is applied to the images of the Set14 dataset \cite{zeyde2010single}.
\figurename~\ref{fig:universal_transfer_set14} shows the super-resolved (SR) images obtained by the RCAN model \cite{zhang2018image}, where the perturbation shown in \figurename~6b of the main paper is applied.
This result verifies that the universal attack is transferable to unseen images.
\\[-0.7\baselineskip]

\begin{figure*}[t]
	\begin{center}
		\centering
		\begin{minipage}[b]{0.20\linewidth}
			\centering
			\centerline{\includegraphics[width=0.98\linewidth]{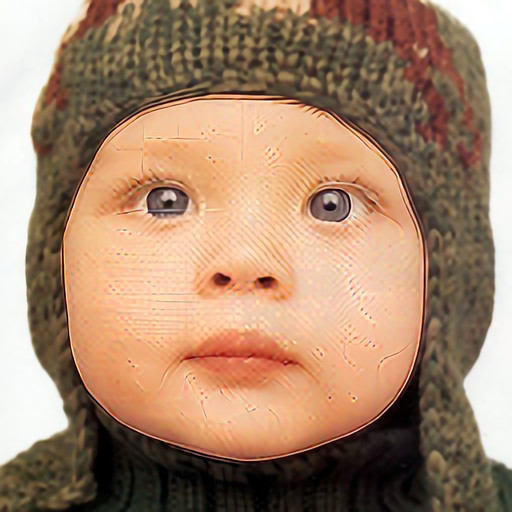}}
			\centerline{EDSR}
		\end{minipage}
		\begin{minipage}[b]{0.20\linewidth}
			\centering
			\centerline{\includegraphics[width=0.98\linewidth]{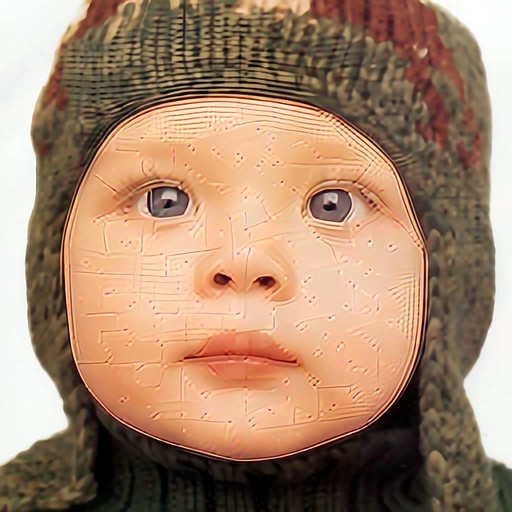}}
			\centerline{RCAN}
		\end{minipage}
		\begin{minipage}[b]{0.20\linewidth}
			\centering
			\centerline{\includegraphics[width=0.98\linewidth]{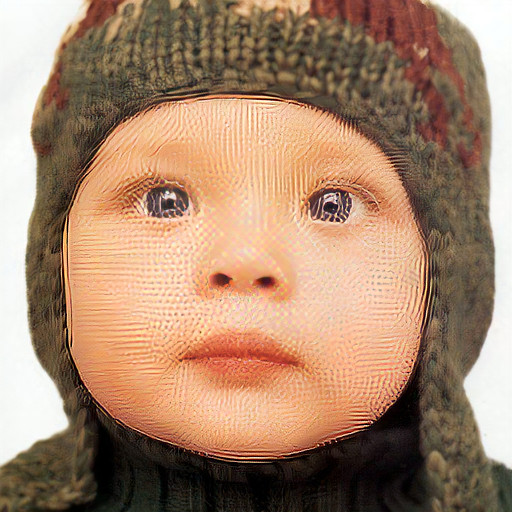}}
			\centerline{4PP-EUSR}
		\end{minipage}
		\begin{minipage}[b]{0.20\linewidth}
			\centering
			\centerline{\includegraphics[width=0.98\linewidth]{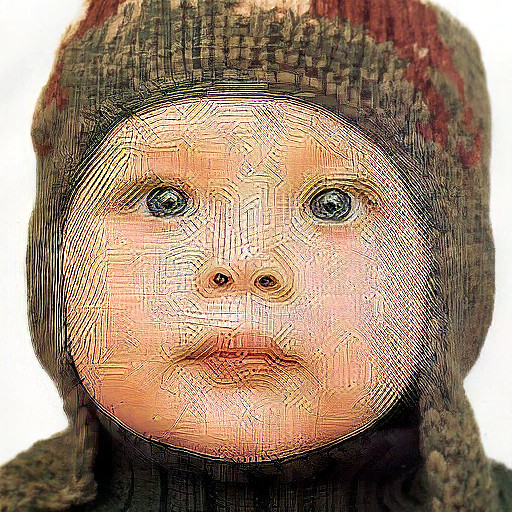}}
			\centerline{ESRGAN}
		\end{minipage}
	\end{center}
	\caption{Results of the partial attack on the face region ($\alpha=16/255$).}
	\label{fig:partial_facemask}
	\vspace{16pt}
\end{figure*}

\noindent \textbf{Advanced partial attack.}
The objective of the partial attack in Section~4.3 of the main paper is to examine how the perturbation planted in a region propagates spatially outside the region.
Partial attacks with more complex masks can also be done using the proposed method.
\figurename~\ref{fig:partial_facemask} shows the attack results where the perturbation is applied on the face region of an image in Set5 \cite{bevilacqua2012low}.
It is observed that strong degradations are introduced around the face boundaries.
\\[-0.7\baselineskip]

\begin{figure*}[t]
	\begin{center}
		\centering
		\begin{minipage}[b]{0.25\linewidth}
			\centering
			\centerline{\includegraphics[width=1.0\linewidth]{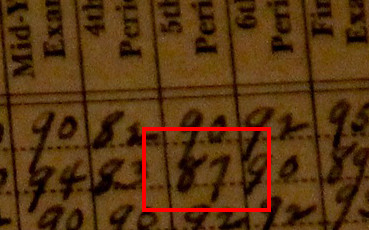}}
			\centerline{HR (original)}
		\end{minipage}
		\begin{minipage}[b]{0.25\linewidth}
			\centering
			\centerline{\includegraphics[width=1.0\linewidth]{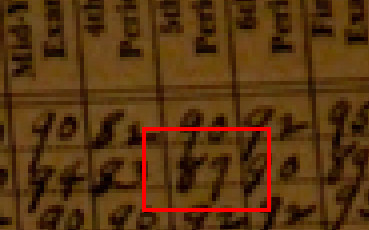}}
			\centerline{LR (attacked)}
		\end{minipage}
		\begin{minipage}[b]{0.25\linewidth}
			\centering
			\centerline{\includegraphics[width=1.0\linewidth]{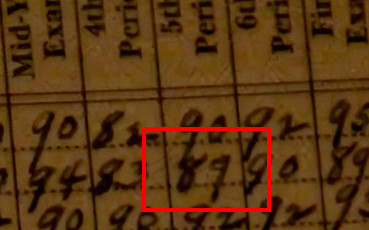}}
			\centerline{SR (attacked)}
		\end{minipage}
	\end{center}
	\caption{Targeted attack result using a score card image (Flickr, juggernautco, CC BY 2.0) with $\alpha=16/255$ for ESRGAN. The attack targets to change the number in the red box to \textit{89}.}
	\label{fig:targeted_record_card}
	\vspace{16pt}
\end{figure*}

\noindent \textbf{Additional example of the targeted attack.}
We provide an additional example of the targeted attack, which is explained in Section 5.1 of the main paper.
\figurename~\ref{fig:targeted_record_card} shows the result.
In the figure, the original number \textit{87} in the original high-resolution image (``HR (original)'') is changed to \textit{89} in the SR version (``SR (attacked)'').
We conduct a subjective test with 20 human observers, and all the observers recognized the number in the red box of ``SR (attacked)'' as \textit{89} instead of \textit{87}.
%This result supports that the adversarial attack with a target image can mislead the super-resolution model in examining the output image of the model.
\\[-0.7\baselineskip]

\begin{figure*}[t]
	\begin{center}
		\centering
		\begin{minipage}[b]{0.48\linewidth}
			\centering
			\centerline{\includegraphics[width=0.98\linewidth]{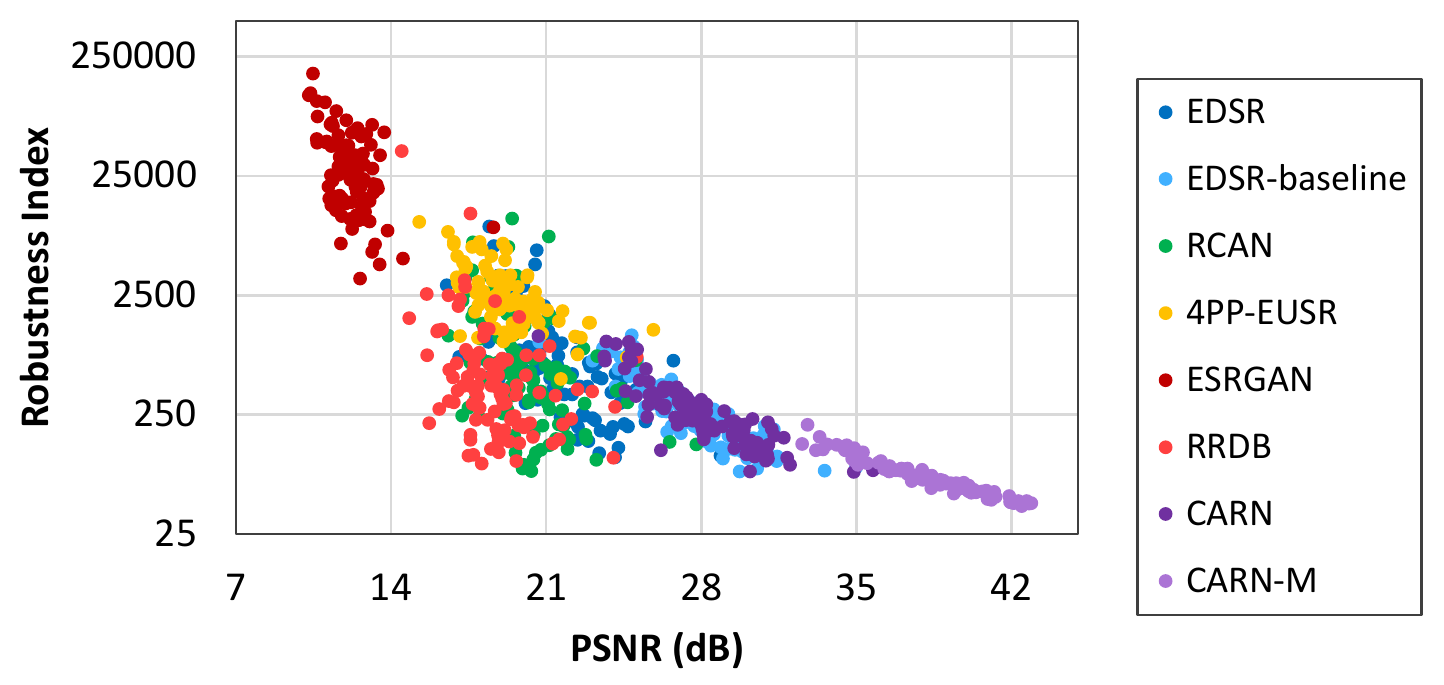}}
			\centerline{$\alpha=2/255$}
		\end{minipage}
		\begin{minipage}[b]{0.48\linewidth}
			\centering
			\centerline{\includegraphics[width=0.98\linewidth]{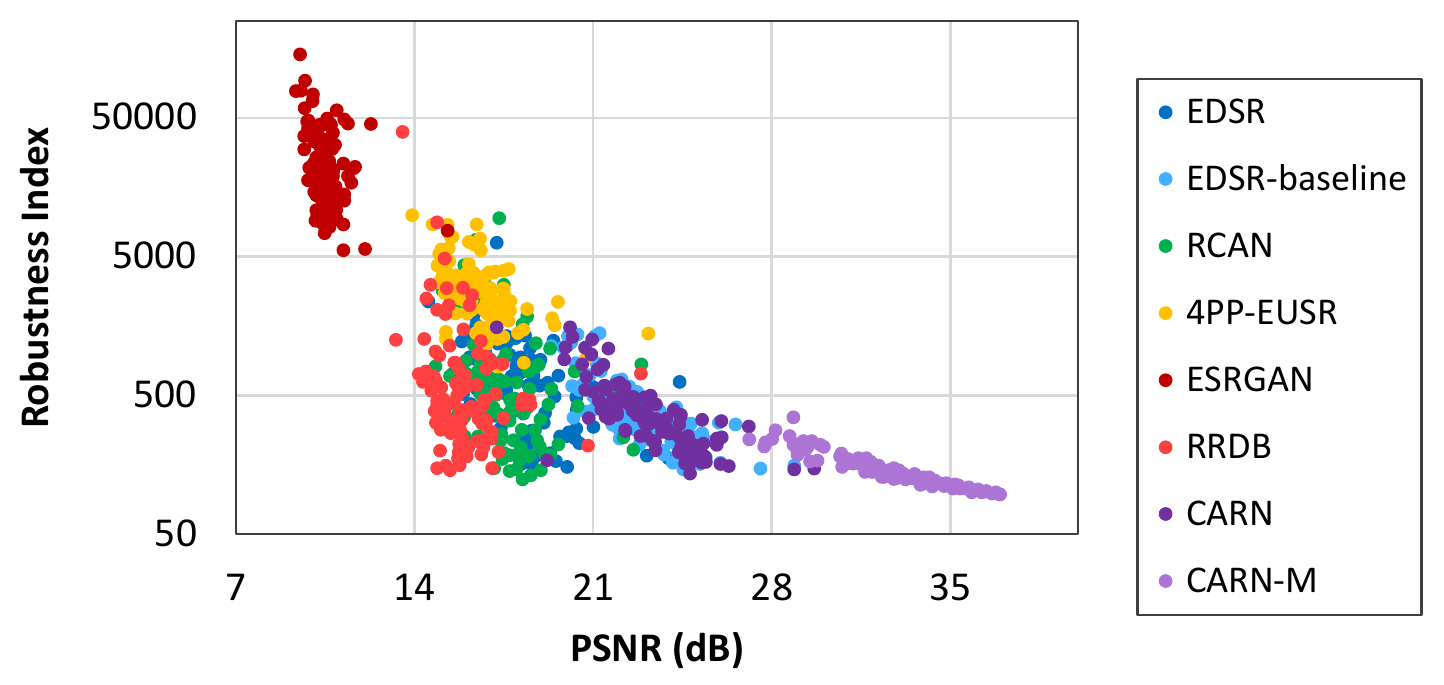}}
			\centerline{$\alpha=4/255$}
		\end{minipage}
	\end{center}
	\caption{PSNR vs. the robustness index for the BSD100 dataset \cite{martin2001database} when $\alpha=2/255$ and $\alpha=4/255$. Each point corresponds to each image in the dataset.}
	\label{fig:psnr_robustness_a2_a4}
\end{figure*}

\noindent \textbf{Robustness measure.}
We employed the ``robustness index'' in Section 5.2 of the main paper.
Here we provide additional results obtained with different $\alpha$ values (i.e., $\alpha=2/255$ and $\alpha=4/255$).
Figure~\ref{fig:psnr_robustness_a2_a4} depicts the relationship between the PSNR values for SR images obtained with the basic attack (Section 4.1 of the main paper) and the robustness indices of the deep super-resolution models for the BSD100 dataset, where $\alpha=2/255$ and $\alpha=4/255$.
When these figures and Figure~10 of the main paper are compared, increasing $\alpha$ results in decreasing the PSNR values and increasing the robustness index values, as expected.
In addition, as in the result with $\alpha=1/255$ (Figure~10 of the main paper), the robustness index is strongly correlated to PSNR regardless of the value of $\alpha$, which supports the usefulness of the robustness index for explaining the relative vulnerability of the different super-resolution methods.

\end{document}